\documentclass{applemlr}

\usepackage{microtype}
\usepackage{graphicx}
\usepackage{subcaption}
\usepackage{booktabs} 
\usepackage{tikz}
\usepackage{pgfplots}
\usepackage{subcaption}
\pgfplotsset{compat=1.18}
\usepackage{hyperref}

\usepackage{amsmath}
\usepackage{amssymb}
\usepackage{mathtools}
\usepackage{amsthm}
\usepackage{enumitem}

\usepackage{amsmath}
\usepackage{enumerate}
\usepackage{algorithm}
\usepackage{algpseudocode}
\usepackage{amsfonts}
\usepackage{amsthm}
\usepackage{cleveref}
\usepackage{diagbox}
\usepackage{colortbl}
\usepackage{amssymb}
\usepackage{xspace}
\usepackage{wrapfig}
\usepackage{adjustbox}
\usepackage{tabularx}
\usepackage{booktabs}
\usepackage{mathtools}
\usepackage{tikz}
\usepackage{enumitem}
\usepackage{silence}
\usepackage{dsfont}
\usepackage[table]{xcolor}
\usepackage[dvipsnames]{xcolor}
\usepackage{multirow}
\usepackage{makecell}
\usepackage{xfakebold}

\usepackage{amsmath,amsfonts,bm}

\def\eqref#1{equation~\ref{#1}}

\def\1{\bm{1}}

\DeclareMathAlphabet{\mathsfit}{\encodingdefault}{\sfdefault}{m}{sl}
\SetMathAlphabet{\mathsfit}{bold}{\encodingdefault}{\sfdefault}{bx}{n}

\definecolor{textgray}{HTML}{6E6E73}
\usetikzlibrary{positioning, calc}
\usetikzlibrary{decorations.pathmorphing}

\makeatletter
\patchcmd{\wrong@fontshape}{\@gobbletwo}{}{}{}
\makeatother
\numberwithin{equation}{section}
\makeatletter
\AtBeginDocument{
  \urlstyle{sf}
  
}
\makeatother

\definecolor{light}{RGB}{125, 125, 125}
\crefname{tcb@cnt@pbox}{code}{code}
\Crefname{tcb@cnt@pbox}{Code}{Code}
\crefname{assumption}{assumption}{assumption}
\Crefname{assumption}{Assumption}{Assumptions}

\newtcolorbox[auto counter]{pbox}[2][]{
  colback=white,
  title=Code~\thetcbcounter: #2,
  #1,fonttitle=\sffamily,
  fontupper=\sffamily,
  arc=2pt,
  colframe=bgcolor,
  coltitle=fgcolor,
  colbacktitle=bgcolor,
  toptitle=0.25cm,
  bottomtitle=0.125cm
}

\makeatletter
\newcommand\applefootnote[1]{%
  \begingroup
  \renewcommand\thefootnote{}%
  \renewcommand\@makefntext[1]{\noindent##1}%
  \footnote{#1}%
  \addtocounter{footnote}{-1}%
  \endgroup
}
\makeatother

\definecolor{cverbbg}{gray}{0.90}

\title{Optimal Splitting of Language Models from Mixtures to Specialized Domains}

\author[*]{Skyler Seto}
\author[*]{Pierre Ablin}
\author[*]{Anastasiia Filippova}
\author[*\dag]{Jiayuan Ye}
\author[*]{Louis Bethune}
\author[*]{Angelos Katharopoulos}
\author[*]{David Grangier}

\affiliation[*]{Apple}
\affiliation[\dag]{National University of Singapore}

\abstract{
    Language models achieve impressive performance on a variety of knowledge, language, and reasoning tasks due to the scale and diversity of pretraining data available.  The standard training recipe is a two-stage paradigm: pretraining first on the full corpus of data followed by specialization on a  subset of high quality, specialized data from the full corpus. In the multi-domain setting, this involves continued pretraining of multiple models on each specialized domain, referred to as split model training.  We propose a method for pretraining multiple models independently over a general pretraining corpus, and determining the optimal compute allocation between pretraining and continued pretraining using scaling laws. Our approach accurately predicts the loss of a model of size $N$ with $D$ pretraining and $D'$ specialization tokens, and extrapolates to larger model sizes and number of tokens.  Applied to language model training, our approach improves performance consistently across common sense knowledge and reasoning benchmarks across different model sizes and compute budgets.
}

\metadata[Correspondence]{\sffamily Skyler Seto: \url{sseto@apple.com}}
\date{\sffamily\today}

\begin{document}

\applefootnote{ \textcolor{textgray}{\sffamily Apple and the Apple logo are trademarks of Apple Inc., registered in the U.S. and other countries and regions.}}

\maketitle

\section{Introduction}

Large language models (LLMs) have demonstrated remarkable performance across a wide range of language understanding and knowledge-intensive tasks~\citep{achiam2023gpt,brown2020language,bubeck2023sparks,liu2024deepseek,team2023gemini}.   Their widespread success can be attributed to the scale of high quality pretraining data with large-scale training datasets exceeding trillions of tokens from the web \citep{hojel2025essential,li2024datacomp}.  Many of the best performing models such as Qwen \citep{yang2025qwen3}, Llama 3 \citep{grattafiori2024llama}, Olmo \citep{olmo20242,olmo2025olmo}, and SmolLM \citep{bakouch2025smollm3} are trained for 4-30 trillion tokens, even for the small model sizes ($<7$B parameters), greatly exceeding the compute-optimal training times recommended by the Chinchilla scaling law \citep{hoffmann2022training}. 

The data used for pretraining (PT) language models is typically derived from large web crawls, and often spans multiple domains of interest; we denote $K$ the number of domains. The goal is to train a model that performs well on average across all domains.  In the setting where $K = 1$, the typical recipe is a two-stage training paradigm: initial pretraining on the full corpus followed by a specialization phase - continued pretraining (CPT) -  on data from the target domain of interest. There are many specialized domains one can choose such as high quality math, code, reasoning, instruction, or multilinguality \citep{bakouch2025smollm3}.  
In many cases, only a short amount of the training time ($20\%)$ is spent on specialization \citep{bakouch2025smollm3,olmo2025olmo}.   

When targeting $K>1$ domains, this strategy can be extended by employing a shared generic pretraining phase followed by independent specialization for each domain. 
In that case, the allocation of the total computational budget between generic pretraining and domain-specific specialization presents a non-trivial optimization problem. Each step in the shared pretraining phase benefits all downstream domains when learning shared features - for example general language syntax and grammars. Conversely, each specialization step benefits only a single domain but may be more effective, as such a step adapts the parameters without the optimization constraints imposed by the other domains - for example when learning fine-grained vocabularies particular to certain field such as science, mathematics, or code.  This is unlike the $K=1$ setting where one should train on the target domain for as long as data is available.

How much of the compute to dedicate to the shared pretraining phase vs. the specialization phase is understudied. Prior works that study CPT don't consider the cost of pretraining in the compute allocation and focus primarily on predicting the loss on the CPT data domain \citep{bethune2025scaling,que2024d,hernandez2021scaling}. Other works show that overtraining in the pretraining phase leads to worse results after finetuning \citep{springerovertrained}, but don't aim to quantify the optimal allocation of compute between pretraining and finetuning, nor do they target multiple domains.

\begin{figure*}[t]
      \centering
      \begin{subfigure}{0.65\textwidth}
          \centering
          \includegraphics[width=\textwidth]{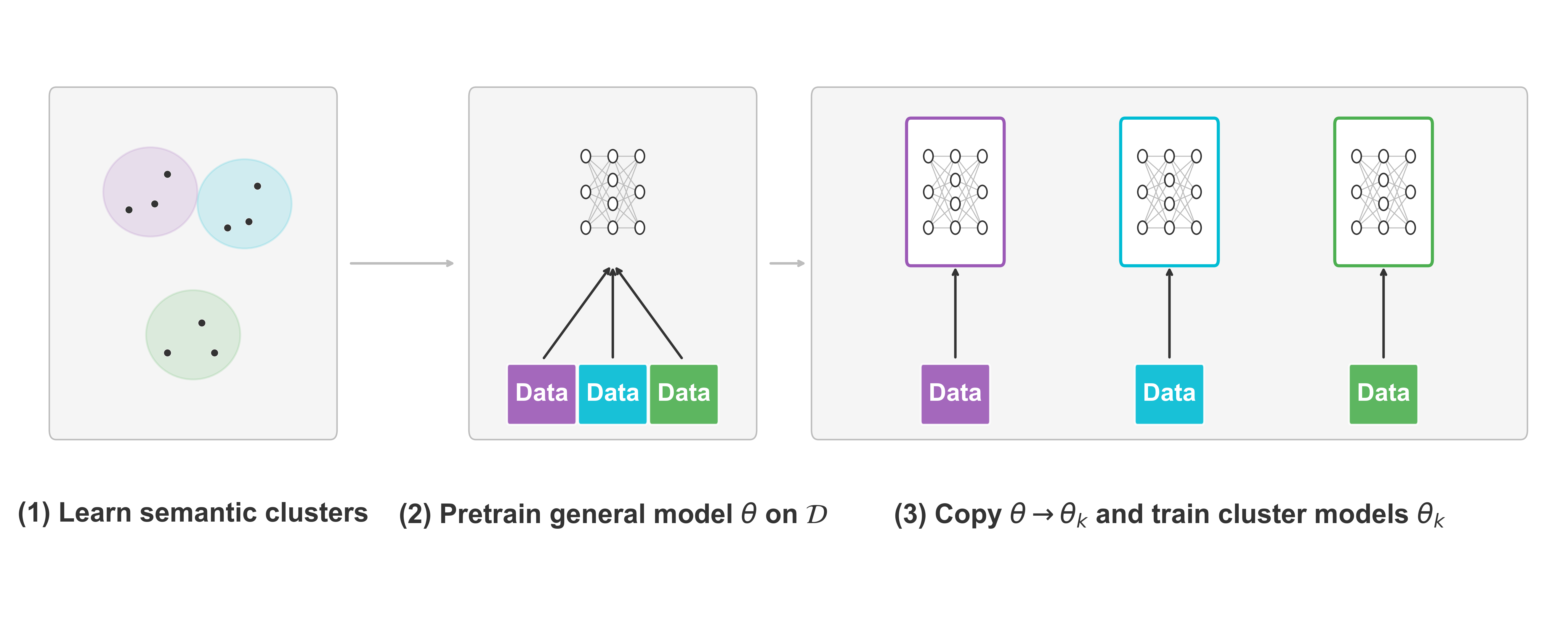}
          \caption{Split model training}
          \label{fig:split_overview}
      \end{subfigure}
      \hfill
      \begin{subfigure}{0.32\textwidth}
          \centering
          \includegraphics[width=\textwidth]{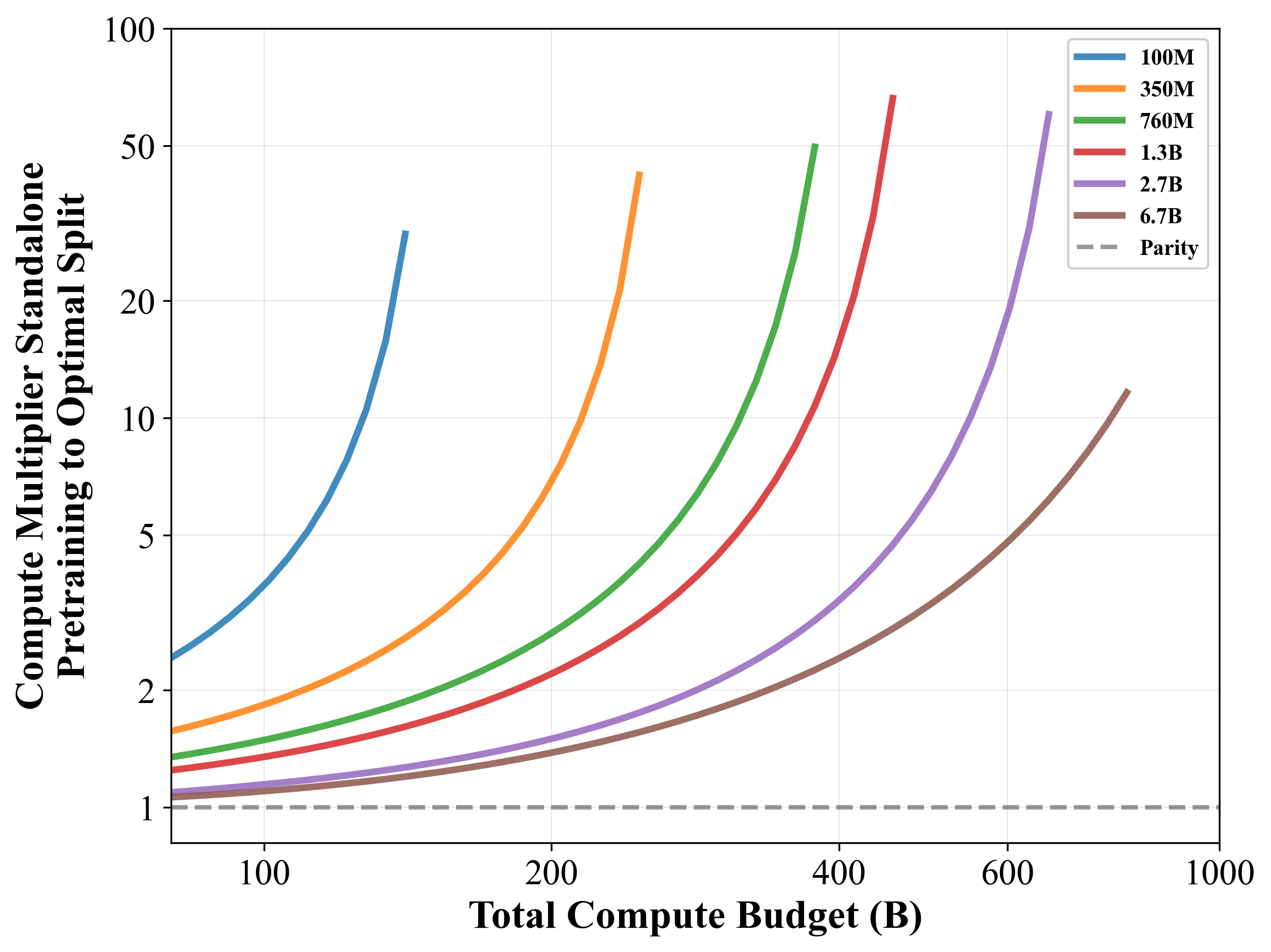}
          \caption{Compute multiplier}
          \label{fig:compute_multiplier}
      \end{subfigure}
      \caption{(a) Split model training: The pretraining corpus is clustered based on semantics, then a model is trained on all domains before being copied and subsequently trained independently on an
  individual cluster. (b) Compute multiplier: The multiplicative factor on the amount of data required for standard pretraining to match optimal split model training performance. At large compute budgets, standard pretraining requires up to
  50$\times$ the amount of data to match the loss of optimal split training.}
      \label{fig:main_fig}
  \end{figure*}

This work explores the question \emph{what is the optimal allocation of compute for pretraining and continued pretraining over multiple data domains?}  We study the scaling behaviors of split model training as a function of pretraining, and specialization data budgets with the goal of demonstrating that split model training with only a small amount of joint pretraining is sufficient.   Our main contributions are  
\begin{itemize}[leftmargin=*]
    \item providing a recipe for pretraining and continued pretraining over multiple domains summarized in \cref{fig:split_overview},
    \item deriving a scaling law that predicts the loss of split model training as a function of compute and identifying the optimal splitting time,
    \item demonstrating that split model training at the optimal compute allocation determined by our scaling law improves over full pretraining in data efficiency in \cref{fig:compute_multiplier}.
\end{itemize}

\section{Related Work}
\textbf{Scaling Laws for Language Models:} Scaling laws aim to predict a model's performance as a function of the total training compute.  Early studies show that the loss of language models exhibit a power law scaling \citep{hoffmann2022training,kaplan2020scaling}. These laws have limitations as they consider only pretraining, a fixed data distribution, and dense models.  Recent studies have expanded the scaling laws to tackle data mixture dependence including optimal data mixture weights \citep{shukor2025scaling,ye2024data}, selection ratios for repetitions of high quality data \citep{goyal2024scaling} and repetitions of data more broadly \citep{muennighoff2023scaling}, and data quality \citep{chang2024scaling}.  Other works  study scaling laws for continued pretraining \citep{que2024d}, and finetuning \citep{bethune2025scaling,zhang2024scaling}, and have been used to measure the effective amount of data from pretraining for code and text tasks \citep{hernandez2021scaling}. Concurrent to our work, scaling laws for bootstrapped continued pretraining are proposed for continued pretraining on a new domain or model growth \citep{liew2025acceleration}. Finally there is a set of work investigating scaling laws for sparsity \citep{abnar2025parameters,krajewski2024scaling,wang2024scaling}, and routing \citep{clark2022unified} in Mixture of Expert (MoE) models.  However, these works do not consider training MoEs with CPT.

\textbf{Split  Model Training:} Prior works have explored split training of models.  One set of these approaches are based on a branch, train, and merge strategy where a base seed model is trained, and individual expert models are trained from the base seed model before merging \citep{li2022branch}.  \citet{gururangan2023scaling} cluster data into different domains and train individual models on each domain before merging, while others train an MoE architecture asynchronously \citep{filippova2025no,shiflexolmo}. These approaches focus on training a mixture and merging asynchronously rather than completely split and routed models, and don't consider pretraining cost as in this work.  Alternatively, several works use several seed models and train an orchestrator that routes queries or parts of a query to different models with limited \citep{dannprincipled,lee2024orchestrallm,wang2025mixllm}, or no additional training \citep{wu2025efficient}, and sometimes to models of varying sizes \citep{chuang2024learning}.  In both settings, a majority of the training is done with a single model limiting efficiency gains from training multiple models, and other benefits of split model training.  

\textbf{Data Mixtures:} Training on a mixture of data domains has become a standard for training language models. Early datasets such as the Pile \citep{gao2020pile}, SlimPajama \citep{cerebras2023slimpajama}, and Dolma \citep{soldaini2024dolma} were aggregated from multiple sources from the web, and recent models are also trained on mixtures of domains \citep{bakouch2025smollm3,olmo2025olmo}.  Other works discover domains through clustering the pretraining corpus \citep{gururangan2023scaling,diaonemotron} extending up to millions of domains \citep{grangier2025task}, or by synthetically generating different data mixtures \citep{maini2024rephrasing,su2025nemotron}.  Our clustering approach is similar to that of \citep{gururangan2023scaling} but we focus on optimal pretraining compute allocation rather than learning to branch and merge split models from an already trained seed model.

\section{Split Model Training}

\label{sec:synthetic}
\subsection{Learning Split Models with Clustering}
Our approach works in three stages, considering both the pretraining (PT) and continued pretraining (CPT) cost, and is highlighted in \cref{fig:split_overview}.  

First, we cluster all documents in the pretraining corpus $\mathcal{D}$ into $K$ disjoint clusters $\mathcal{D}_1, \mathcal{D}_2, \dots, \mathcal{D}_K$ using document embeddings from a document embedding model.  A router $\mathcal{R}(x; W)$ is learned from the clustering to map from documents to cluster assignments.
Second, we perform a two stage training of $\theta$.  A seed model $\theta'$ is trained on $D$ tokens from $\mathcal{D}$ to approximately minimize
\begin{equation}
\theta(D) = \arg\min_{\theta} \; \mathbb{E}_{x \sim \mathcal{D}}\big[ \mathcal{L}(x;\theta) \big],
\end{equation}
with the next-token prediction loss
\begin{equation}
    \mathcal{L}(x; \theta) = -\sum_s \log p_{\theta}(x_s | x_{<s}).
    \label{eq:main_loss}
\end{equation}
After training $\theta(D)$, we create $K$ copies $\theta_1, \theta_2, \dots, \theta_K$ and train each on $\mathcal{D}_1, \mathcal{D}_2, \dots, \mathcal{D}_K$ respectively with $D'$ tokens
\begin{equation}
\theta_k  = \arg\min_{\theta} \; \mathbb{E}_{x\sim \mathcal{D}_k}\left[\mathcal{L}(x; \theta)\right].
\end{equation}
We refer to this as \emph{split model training} as all of the data is split across $K$ models which are trained completely independently from the seed model.  The total training cost of split model training is a function of the total number of tokens $D + K \times D'$, where $D$ is the total number of pretraining tokens, and $D'$ is the number of tokens each expert model is trained on.

At inference, given a new document $x$, a prefix $x_p$ (such as a question, starting sentence, or context passage) is selected, and the router selects the corresponding cluster index $c$ as
\begin{equation}
    c = \arg\min_{c} \mathcal{R}(x_p; W) = \|x_p - W_c\|.
\end{equation}
The model $\theta_c$ is then used on the entire document $x$.  

Prior works that train split models use a pretrained seed model and do not consider the amount of data for PT  $D'$ and CPT $D'$ \citep{li2022branch,gururangan2023scaling}.  These works train on the order of hundreds of billions of tokens, but a large part ($>50\%$) of the training is still done in the pretraining phase. They do not consider whether a short or no pretraining phase could yield similar or better performance. We start by demonstrating that this is an important consideration as pretraining for too short or too long can reduce performance under the same total training budget, and that the optimal time to split depends on the per-cluster data distributions and the total compute budget.  



\subsection{Phonebook Memorization Experiments}
\textbf{Implementation Details:} We first motivate that split model training has varying performance improvements compared with pretraining on all domains. We start with an experiment on fact memorization in transformers.  For our experiments, we train a transformer model with 12.9M parameters (6 layers, 8 attention heads, and hidden dimension 256). We fix the learning rate at $0.0001$, batch size as 640, and warmup steps as 20000.  We additionally study perplexity improvements when training 1.3B parameter language models on three pairs of clusters from a clustered DCLM dataset in Appendix~\ref{sec:two_cluster}.

\textbf{Methodology:} We follow \citep{jelassi2024mixture} and use synthetically generated phonebook records to measure fact memorization.  Each cluster contains 80000 randomly generated phonebook facts, where each fact is a (name, phone-number) tuple of the format 
$$<6 \text{ alphabet tokens}>|<22 \text{ digit tokens}>,$$
where the name contains six randomly drawn alphabetical tokens from a to z, and the phone-number contains 22 randomly sampled digits from 0 to 9. In such setting, we define similarity between the two clusters as $\alpha = \frac{|\text{cluster}_1 \cap \text{ cluster}_2|}{80000}$.  

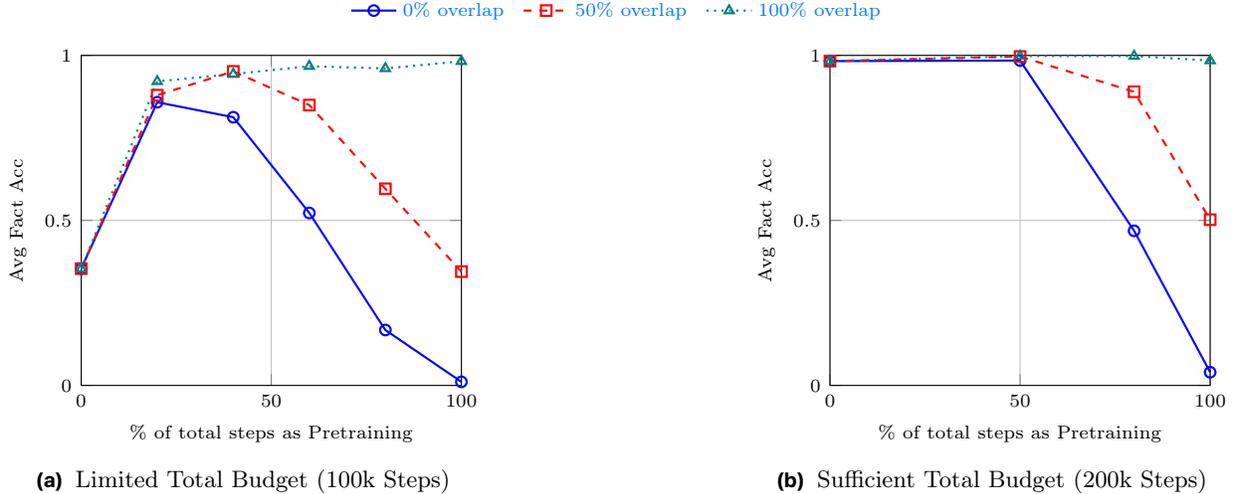
\begin{figure}[t]
    \centering
    \ref{legend:phonebook_cpt}\\[0.3em]
    \begin{subfigure}[b]{0.4\columnwidth}
        \centering
        \begin{tikzpicture}
            \begin{axis}[
                width=\textwidth,
                height=0.9\textwidth,
                xlabel={\% of total steps as Pretraining},
                ylabel={Avg Fact Acc},
                xmin=0, xmax=100,
                ymin=0, ymax=1,
                xtick={0, 50, 100},
                ytick={0, 0.5, 1.0},
                grid=major,
                mark size=2pt,
                tick label style={font=\scriptsize},
                label style={font=\scriptsize},
            ]
            \addplot[blue, thick, mark=o, mark options={solid, fill=blue}, solid] coordinates {
                (0, 0.353775)
                (20, 0.8580875)
                (40, 0.812675)
                (60, 0.5225)
                (80, 0.16785)
                (100, 0.01078125)
            };
            \addplot[red, thick, mark=square, mark options={solid, fill=red}, dashed] coordinates {
                (0, 0.353775)
                (20, 0.8794)
                (40, 0.9515125)
                (60, 0.8494375)
                (80, 0.5957875)
                (100, 0.3447583)
            };
            \addplot[teal, thick, mark=triangle, mark options={solid, fill=teal}, dotted] coordinates {
                (0, 0.353775)
                (20, 0.9211125)
                (40, 0.9434625)
                (60, 0.9672125)
                (80, 0.96045)
                (100, 0.9821)
            };
            \end{axis}
        \end{tikzpicture}
        \caption{Limited Total Budget (100k Steps)}
        \label{fig:phonebook_cpt_short_overlap}
    \end{subfigure}
    \hfill
    \begin{subfigure}[b]{0.4\columnwidth}
        \centering
        \begin{tikzpicture}
            \begin{axis}[
                width=\textwidth,
                height=0.9\textwidth,
                xlabel={\% of total steps as Pretraining},
                ylabel={Avg Fact Acc},
                xmin=0, xmax=100,
                ymin=0, ymax=1,
                xtick={0, 50, 100},
                ytick={0, 0.5, 1.0},
                grid=major,
                mark size=2pt,
                tick label style={font=\scriptsize},
                label style={font=\scriptsize},
                legend to name=legend:phonebook_cpt,
                legend columns=3,
                legend style={
                    draw=none,
                    font=\scriptsize,
                    /tikz/every even column/.append style={column sep=0.5em},
                },
            ]
            \addplot[blue, thick, mark=o, mark options={solid, fill=blue}, solid] coordinates {
                (0, 0.9826625)
                (50, 0.98435)
                (80, 0.468425)
                (100, 0.0399125)
            };
            \addlegendentry{0\% overlap}
            \addplot[red, thick, mark=square, mark options={solid, fill=red}, dashed] coordinates {
                (0, 0.9826625)
                (50, 0.996575)
                (80, 0.889775)
                (100, 0.50215)
            };
            \addlegendentry{50\% overlap}
            \addplot[teal, thick, mark=triangle, mark options={solid, fill=teal}, dotted] coordinates {
                (0, 0.9826625)
                (50, 0.99805)
                (80, 0.9979625)
                (100, 0.98435)
            };
            \addlegendentry{100\% overlap}
            \end{axis}
        \end{tikzpicture}
        \caption{Sufficient Total Budget (200k Steps)}
        \label{fig:phonebook_cpt_overlap}
    \end{subfigure}
    \caption{Average fact accuracy (over two clusters) vs. split step for synthetic phonebook fact clusters with different overlaps under different total training budget. We say the total budget is sufficient if split training on two cluster each using (Total Steps/2) from step 0 reaches 100\% average fact accuracy, and otherwise say the total budget is limited.}
    \label{fig:phonebook_cpt}
\end{figure}

\textbf{Findings:} Results for 100K steps of training are in \cref{fig:phonebook_cpt_short_overlap} and 200K steps in \cref{fig:phonebook_cpt_overlap}.   In the limited budget setting, the best performing model varies with the degree of overlap.  In the 100\% overlap setting, it is always better to pretrain for longer on both domains.  In the 0\% and 50\% settings, the average fact accuracy peaks when allocating 20\%-40\% of the total compute budget to pretraining, then subsequently decreases zero as we allocate more compute budget to pretraining, as the amount of combined information in two clusters is too large for a single pretrained model to memorize accurately.  In the sufficient budget setting, similar trends are shown, but here the beginning of the curve (early splitting) also achieves perfect memorization.

\section{Scaling Laws for Split Model Training}
In Section~\ref{sec:synthetic}, we found that split model training improvements depend on the amount of PT, CPT, total compute budget, and domain similarity.  
Predicting the final model loss from independent model training as a function of the number of pretraining tokens $D$, and identifying optimal values of $D$, can lead to better-trained models with greater efficiency.  We consider neural scaling laws for predicting the test loss of the model.

\subsection{Setup}

Our prediction of $\mathcal{L}$ is a function of the learning algorithm which has parameters $D_1, \dots, D_K$, the number of tokens for training from each subdomain, and $N$, the number of parameters in $\theta$. Note that predicting the loss of all $k$ domains is equivalent to predicting the loss of individual domains and taking the weighted average.  Thus, for simplicity we consider estimation of the loss $\mathcal{L}(N, D, D_k)$ of domain $k$ from which we can compute the average loss over all domains.

\subsection{Scaling Law Functional Form}
\label{sec:functional_form}
Neural scaling laws describe a model's loss as a function of the model size, amount of training data, and compute budget.  The seminal Chinchilla scaling law for language models is an additive law of the form
\begin{equation}
\label{eq:chinchilla}
    \mathcal{L}(N, D) = E + A\cdot N^{-\alpha} + B\cdot D^{-\beta}
\end{equation}
where $E, A, \alpha, B, \beta$ are learnable parameters dependent on the data, model architecture, and optimization procedure \citep{hoffmann2022training}.  

A key difference in our work is that we consider a two-stage training with $D$ and $D_k$, which leads to differences in the loss functional form. First, the model should be a function of both $D$ and $D_k$, and in the absence of the other, both should follow a power law. That is, training only on $D_k$ and only on $D$ should independently follow a standard Chinchilla scaling law.  Second, we note that as $\mathcal{D}_k \subset \mathcal{D}$, training on $D$ should improve loss on $D_k$ but at a slower rate. 
We have the following desiderata for a scaling law that models $\mathcal{L}(N, D, D_k)$:
i) the limited capacity of the model prevents it from learning all of $\mathcal{D}$ and $\mathcal{D}_k$ from $\mathcal{D}$. Therefore, we would like the scaling law asymptotics to predict a different irreducible loss when pretraining with infinite compute on $D$ compared to training only on  $D_k$,
ii) $\mathcal{L}(N, D, D_k)$ should recover a Chinchilla type-scaling when either $D$ or $D_k = 0$, and
iii) taking $D_k$ to infinity should lead to the same irreducible loss, regardless of $D$.
 In order to satisfy i), we consider a bias $E_{\Delta}$, which is a sigmoid function of $D_k$ and $N$:
\begin{equation}
    \label{eq:pt_bias}
    E_{\Delta} = E_p \cdot \frac{1}{1+(N/N_s)^{\gamma_1}} \cdot \frac{1}{1+(D_k/D_s)^{\gamma_2}}.  
\end{equation}
Our final law is 
\begin{equation}
    \label{eq:scaling_law_with_bias}
    \mathcal{L}(N, D, D_k) = E_0 + E_{\Delta} + A\cdot(D_k^{\alpha_1} + cD^{\alpha_2})^{-1} + B\cdot N^{-\kappa}
\end{equation}
where $E_p$, $E_0$, $N_s$, $D_s$, $A$, $B$, $\gamma_1$, $\gamma_2$, $\alpha_1$, $\alpha_2$, $c$ and $\kappa$  are all learnable parameters. Note that our scaling law handles the conditions stated above with an irreducible loss independent for both $D$ and $D_k$. 

\section{Experiments for Scaling Laws}

\subsection{Experimental Setup}
\label{experiment_setup}
 We use decoder only transformer models in the parameter sizes of 100M, 350M, 760M, 1.3B,  2.7B parameters. All models are trained using the DCLM dataset with a batch size of 1M tokens\footnote{We use a context length of 1024 and 1024 samples per batch.  This is approximately 1M tokens and we use the number of steps in the thousands interchangeably with billions of tokens.} following the hyperparameters in Appendix~\ref{sec:hyperparams}.  The DCLM dataset is clustered using a balanced K-means clustering with 16 clusters following details in Section~\ref{sec:cluster_details}.  
 The scaling law is fitted using the basin-hopping algorithm and the methodology described in~\citep{shukor2025scaling} and in Appendix~\ref{app:scalefit}.


\begin{figure*}[ht]
    \centering
    \includegraphics[width=0.32\textwidth]{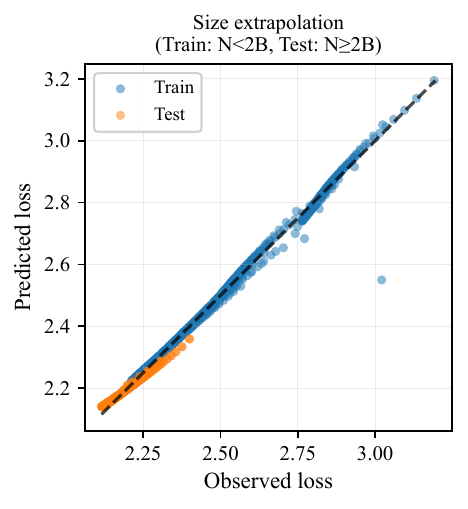}
\includegraphics[width=0.32\textwidth]{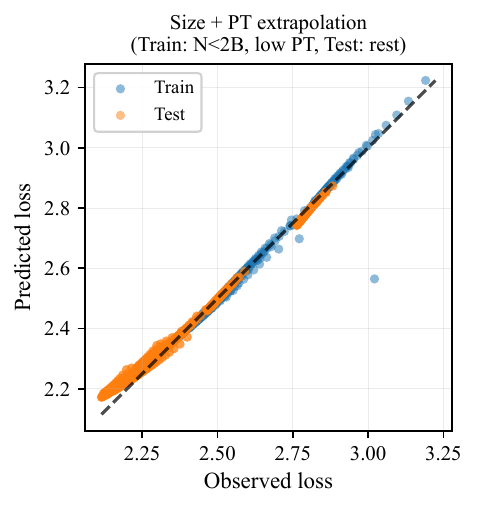}
\includegraphics[width=0.32\textwidth]{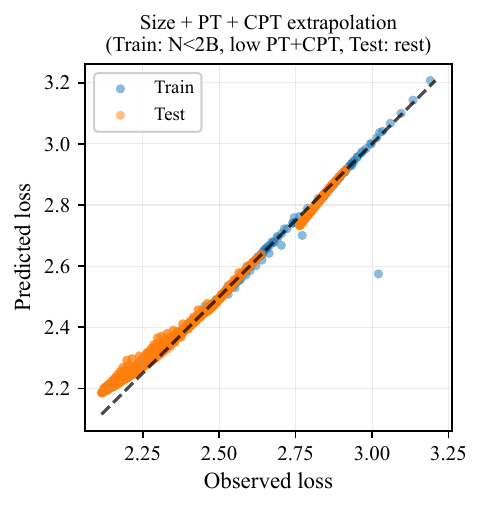}
    \caption{Predicted loss vs. observed loss across multiple scenarios.  The scaling laws are fit on smaller models with varying amount of PT and CPT token budgets.  Loss is estimated over domain 8.}\label{fig:scaling_law_loss_observed_pred}
\end{figure*}

\subsection{Scaling Law Fitting}
\label{sec:scaling_law_fitting}
We train the scaling law in (3) using data of the form $(D, D_k, N, \mathcal{L})$ obtained by training models across a range of values and estimating the loss $\mathcal{L}$ on a validation set of data from $D_k$ with $k=8$ and total number of domains $K=16$.  We consider three scenarios when testing the scaling law:
\begin{itemize}
    \item Scenario 1: Train on small model size ($\leq 1.3$B), and test on larger model size ($2.7$B)
    \item Scenario 2: Train on small model size ($\leq 1.3$B) and lower half of the range of  $D$, and test on rest
    \item Scenario 3:  Train on small model size ($\leq 1.3$B) lower half of the range of $D$ and $D_k$, and test on rest.
\end{itemize}

Table~\ref{tab:scaling_law_mre} describes our results in terms of the mean absolute error (MAE) and $R^2$ coefficient.  Both MAE is low from $0.012-0.023$, and the $R^2$ is high and close to 1 even for setting where we test on only a small handful of runs - 130 for scenario 3. When testing with similar number of runs but for smaller models we have overall low MAE and high $R^2$.

\begin{table}[ht]
    \centering
    \begin{tabular}{|l|c|c|c|c|}
    \hline
         & Train Size & Test Size &  MAE & R2 \\
       \hline
        Scenario 1 & 652 & 144 & 0.012 &  0.934 \\
        Scenario 2 & 347 & 449 & 0.021 & 0.982\\
        Scenario 3 & 156 & 640 & 0.023 & 0.979 \\
        \hline
    \end{tabular}
    \caption{MAE and $R^2$ for different model size on varying held out test conditions.}
    \label{tab:scaling_law_mre}
\end{table}

Additionally, we plot the observed loss vs. predicted loss in \cref{fig:scaling_law_loss_observed_pred}.  We find that the predicted loss and observed loss are extremely close, and that both train and test points do not deviate from the fit.  
A comparison with \citep{liew2025acceleration} is available in Appendix~\ref{app:sec:comparison_liew}. 






\subsection{Optimal Number of Pretraining Tokens}

We now use the previous scaling laws to predict the optimal number of pretraining tokens, after which it becomes worthwhile to specialize and split models.

\begin{figure}[t]
\centering
    \includegraphics[width=0.4\columnwidth]{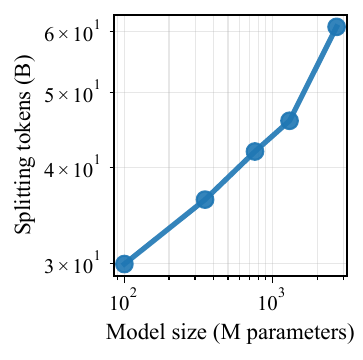}
    \hfill
\includegraphics[width=0.40\columnwidth]{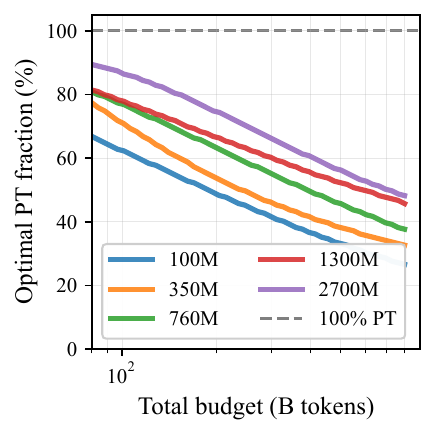}
    \caption{\textbf{Left} Number of pretraining tokens after which split training becomes beneficial $D_{\mathrm{split}}$, as a function of model size.
    We compute it as the solution of \eqref{eq:split_id}.
    We see that it takes more tokens for larger models to start benefiting from split training.
    We only use the loss on one domain to approximate the average loss $L$.
    \textbf{Right} Optimal fraction of pretraining tokens as a function of the total training budget $t_S/ D_T$: the fraction decays with budget, but the optimal number of pretraining tokens increases with budget.}
    \label{fig:splitting_time}
\end{figure}

Taking model size aside, define $\mathcal{L}_k(D, D')$ as the loss of a model pretrained for $D$ tokens on $\mathcal{D}$ and then trained for $D'$ tokens on $\mathcal{D}_k$.
Since training is split, the total number of tokens used to train these $K$ models is $D_T = D + K\times D'$.
Therefore, assuming we are looking for the best models on average over all domains that minimize $L(D, D'):=\sum_{k=1}^K \mathcal{L}_k(D, D')$, the question of the optimal number of pre-training tokens becomes that of optimal allocation:
\begin{equation}
    \min_{D, D'} L(D, D') \text{ s.t. } D + K\times D' = D_T
    \label{eq:allocation}
\end{equation}
We define $t_S$ as the \textbf{optimal splitting point}, that is, the number of pre-training tokens that solves the previous problem. 
It depends on the total budget $D_T$.
First, we may wonder if it is optimal to split training at all. 
Say that we have pretrained a model for $D$ tokens, and that we have an additional infinitesimal budget of $\delta$ tokens. 
We can either a) use these $\delta$ tokens to keep on pretraining on $D$, which will give us a loss $L(D + \delta, 0)\simeq L(D, 0) +  \frac{\partial L(D_{\mathrm{split}}, 0)}{\partial D}\delta$, or b) we can split the models and train on each domain with $\delta / K$ tokens. This will give us a loss
$L(D, \delta / K)\simeq L(D, 0) + \frac1K\frac{\partial L(D_{\mathrm{split}}, 0)}{\partial D'}\delta$.
The \textbf{minimal splitting point} $D_{\mathrm{split}}$ is such that b) becomes more advantageous than a), which gives 
\begin{equation}
K\frac{\partial L}{\partial D}(D_{\mathrm{split}}, 0)= \frac{\partial L}{\partial D'}(D_{\mathrm{split}}, 0)
\label{eq:split_id}
\end{equation}
If the total budget is such that $D_T\leq D_{\mathrm{split}}$, splitting is useless and the optimal allocation is to put $t_s=D_T$.
\cref{fig:splitting_time} shows this splitting point as a function of model size.

\textbf{Findings:} First, we note that as model size increases, the minimal splitting point $D_{\mathrm{split}}$ increases. For small model sizes, the curves appears to be linear in model size.  Second, we empirically remark greedy splitting is not optimal and that the solution to \eqref{eq:allocation} is \emph{not} $t_S = D_{\mathrm{split}}$, i.e., to pretrain for $D_{\mathrm{split}}$ and then do continued pretraining: if we have a larger budget, it might be worthwhile to pretrain for longer than $D_{\mathrm{split}}$, and $t_s\geq D_{\mathrm{split}}$.
We fit the scaling law in ~\eqref{eq:scaling_law_with_bias}, and then we report the optimal number of pretraining tokens for each budget in \cref{fig:splitting_time}. 
This number is not constant past the splitting point $D_{\mathrm{split}}$. 
A greedy splitting solution that splits a model as soon as the budget goes above $D_{\mathrm{split}}$ is therefore not optimal; optimal splitting point must be aware of the training horizon.


\section{Language Model Performance}
\label{sec:exp_lm}
We compare scaling law results with LLM downstream performance on zero-shot question answer performance. We report the average accuracy across eight benchmarks: ARC-Easy, ARC-Challenge, HellaSwag, PIQA, SciQ, BoolQ, MMLU, and Winogrande. For additional details see Appendix~\ref{sec:evaluation_datasets}. We show that QA performance improves with split model training, and that performance matches models with larger inference costs. Next, we apply our optimal split point to train 1.3B and 2.7B parameter models that outperform models with larger inference budget. Finally, we consider some ablations on the number of models, and architecture trained.  

\subsection{Optimal Model Split Point for Varying Budget}
\textbf{Methodology:} We train 1.3B parameter models at varying amounts of compute in approximately \{120, 240, 360\} billion total tokens. We train several models with varying amount of pretraining in \{0, 10, 20, 60, 80, 120, 160, 220, 280, 320, and 340\} billion tokens, and additionally report the optimal $t_s$ as reported by the scaling law. 

\begin{figure*}[t]
    \centering
       \begin{subfigure}{0.3\textwidth}
        \centering
        \includegraphics[width=\textwidth]{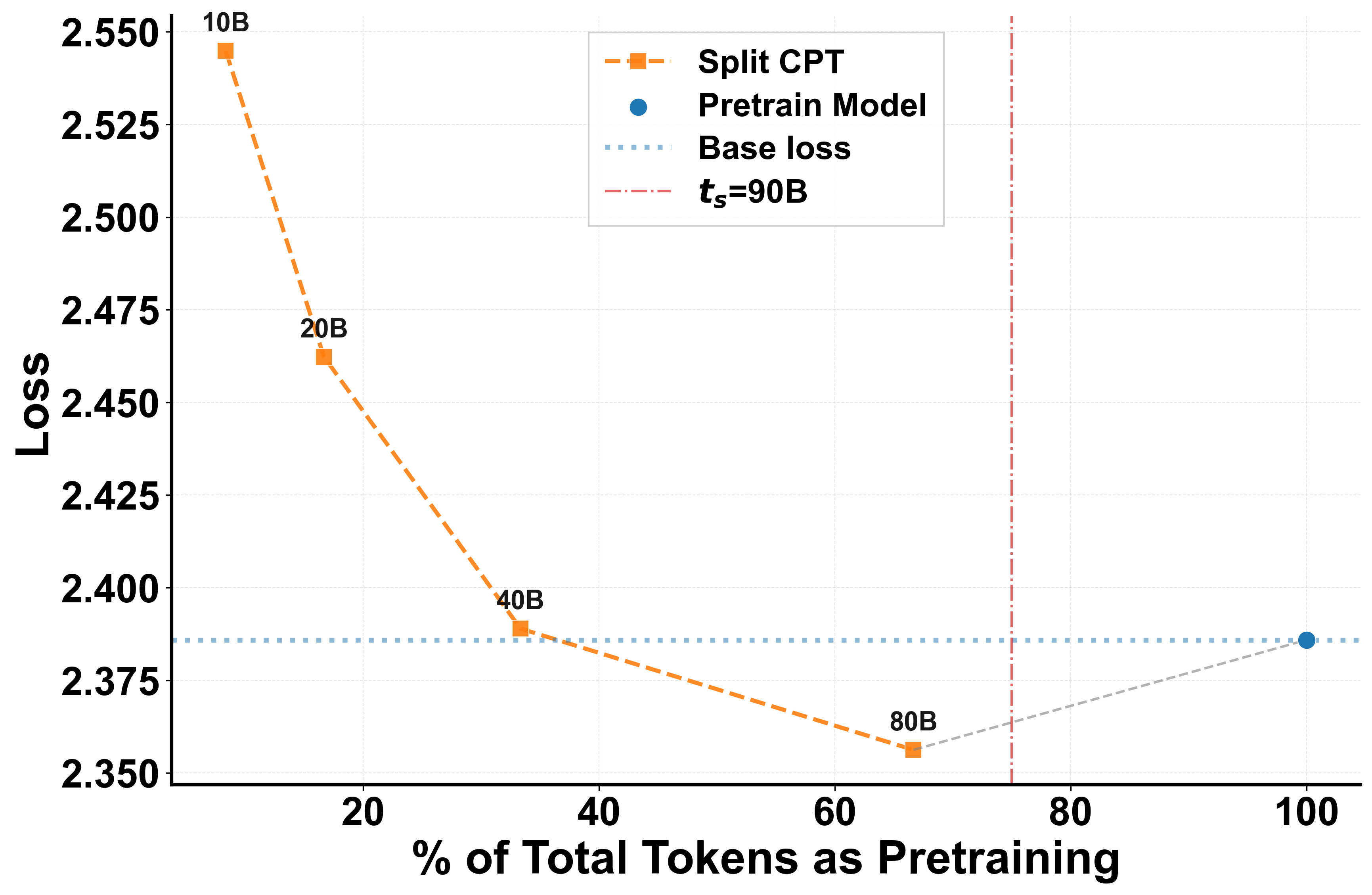}
        \caption{120B}
        \label{fig:120k_pt_loss}
    \end{subfigure}
    \begin{subfigure}{0.3\textwidth}
        \centering
        \includegraphics[width=\textwidth]{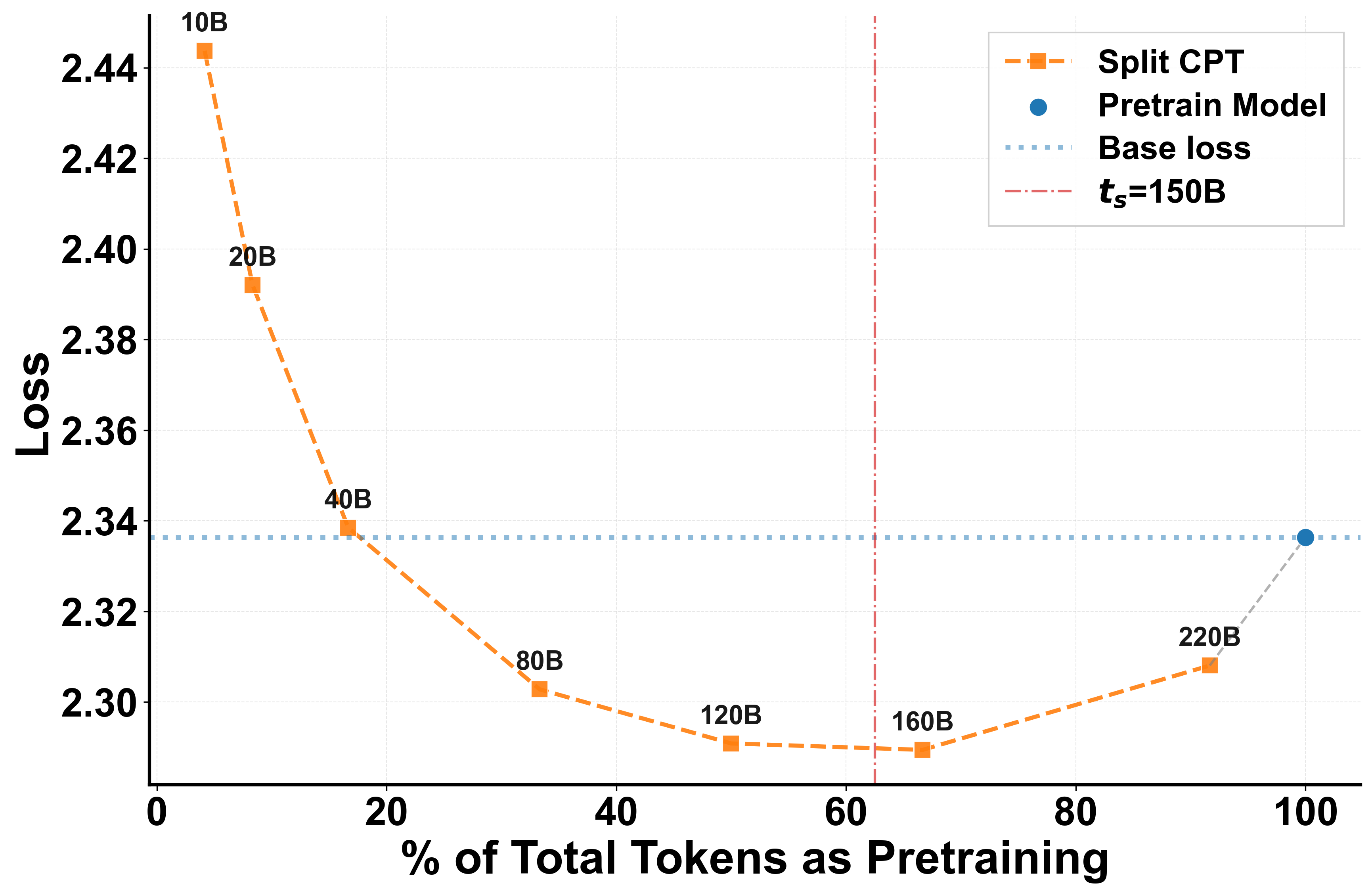}
        \caption{240B}
        \label{fig:240k_pt_loss}
    \end{subfigure}
     \begin{subfigure}{0.3\textwidth}
        \centering
        \includegraphics[width=\textwidth]{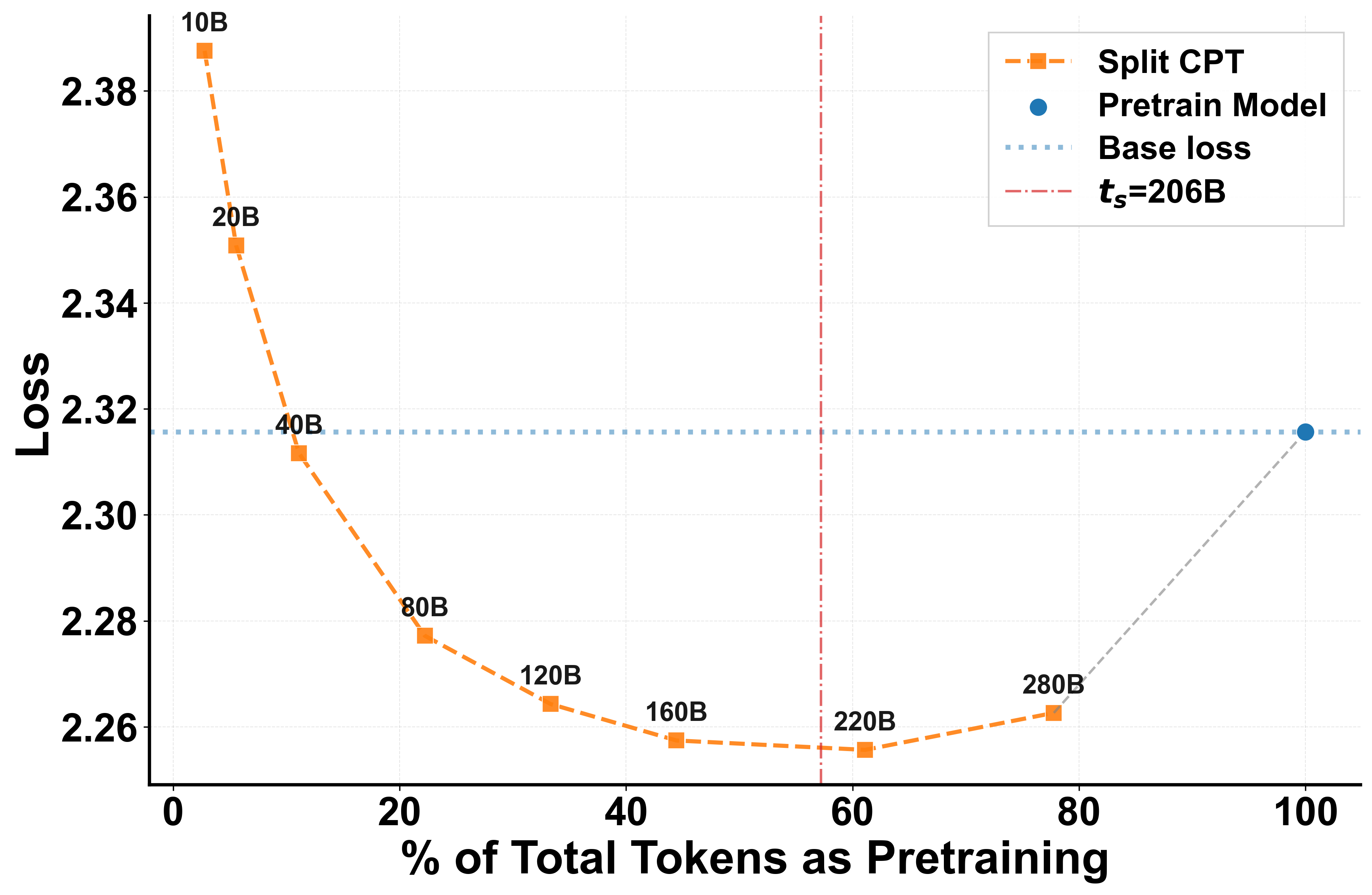}
        \caption{360B}
        \label{fig:300k_pt_loss}
    \end{subfigure}
    \caption{Loss vs. fraction of pretraining tokens for a 1.3B parameter models at 120, 240, and 360B total training tokens.}
    \label{fig:loss_1_3B}
\end{figure*}

\begin{figure*}[t]
    \centering
       \begin{subfigure}{0.3\textwidth}
        \centering
        \includegraphics[width=\textwidth]{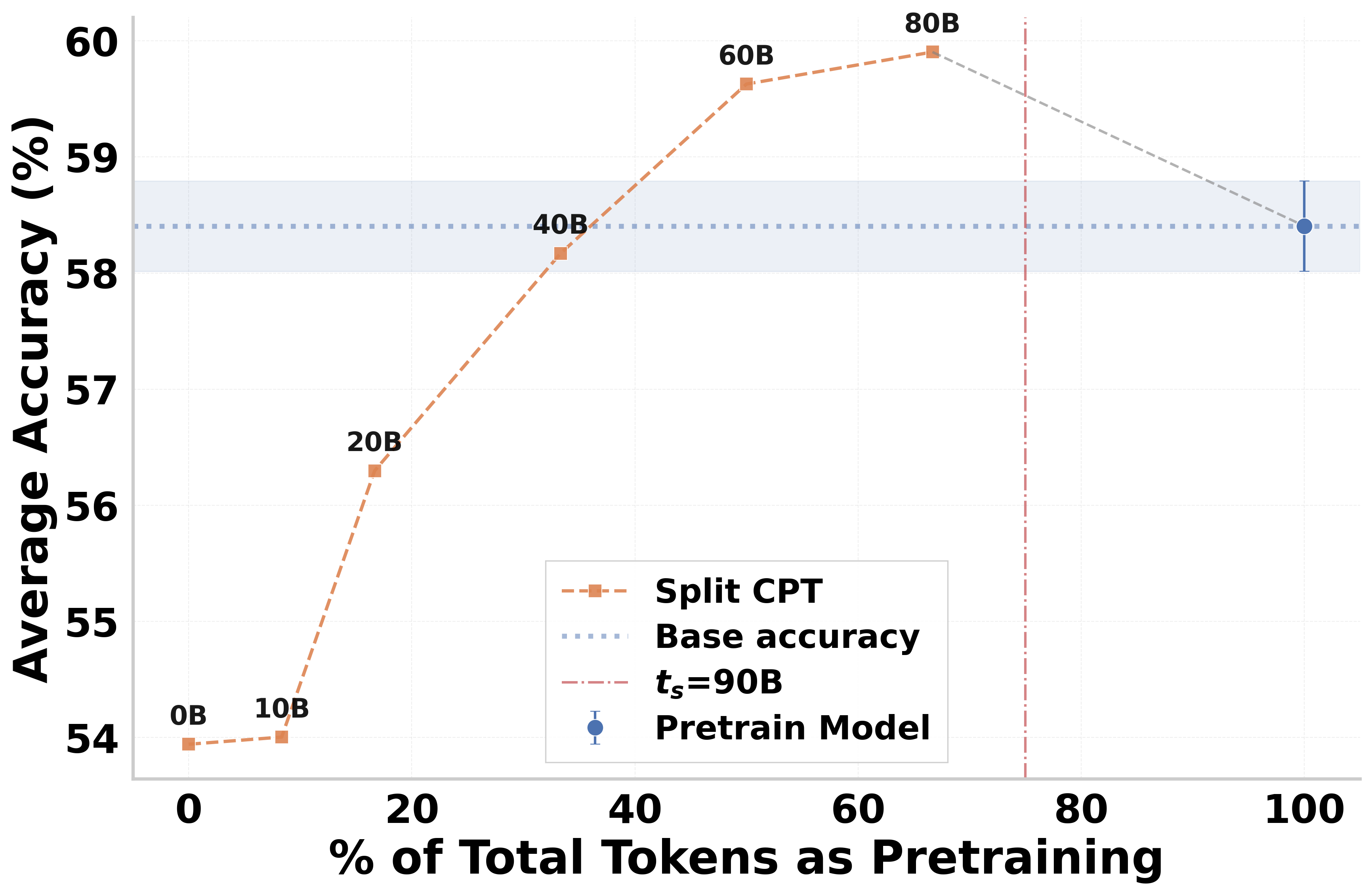}
        \caption{120B}
        \label{fig:120k_pt}
    \end{subfigure}
    \begin{subfigure}{0.3\textwidth}
        \centering
        \includegraphics[width=\textwidth]{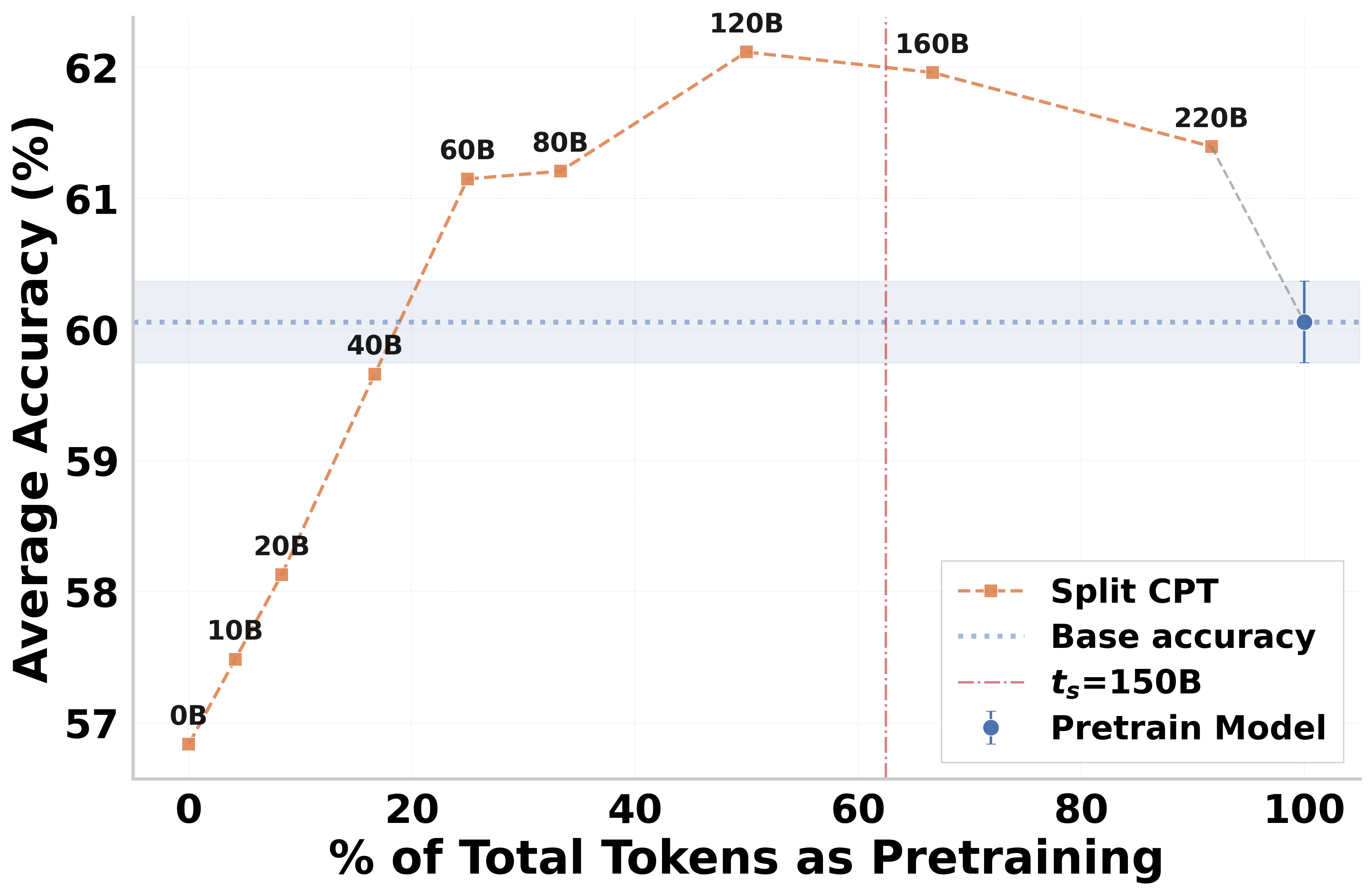}
        \caption{240B}
        \label{fig:240k_pt}
    \end{subfigure}
     \begin{subfigure}{0.3\textwidth}
        \centering
        \includegraphics[width=\textwidth]{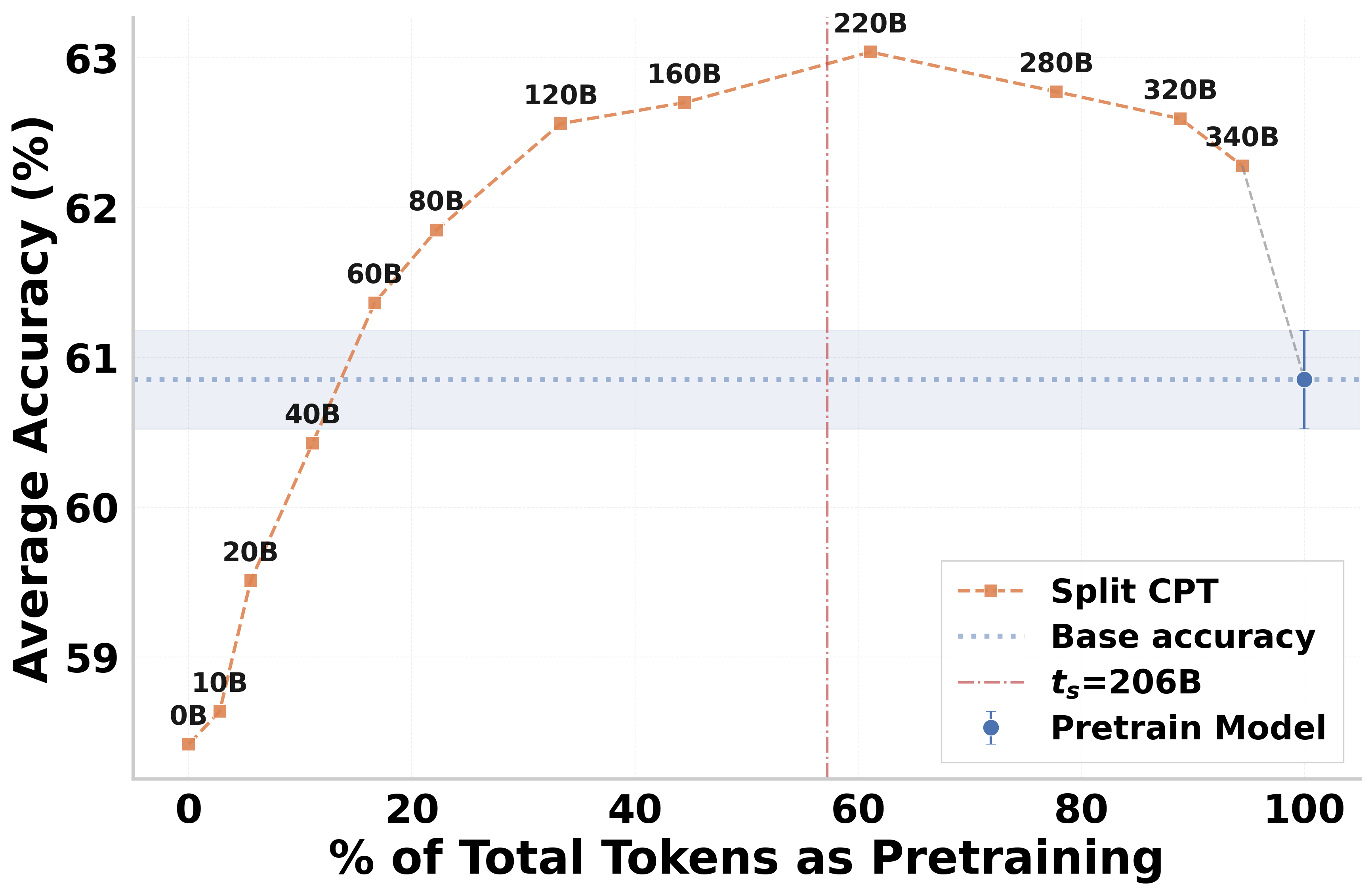}
        \caption{360B}
        \label{fig:300k_pt}
    \end{subfigure}
    \caption{Zero shot QA accuracy vs. fraction of pretraining tokens for a 1.3B parameter models at 120, 240, and 360B total training tokens. }
    \label{fig:zero_shot_qa_1_3B}
\end{figure*}

\textbf{Findings:} We present results in \cref{fig:loss_1_3B} for loss and \cref{fig:zero_shot_qa_1_3B} for zero-shot QA benchmarks. For all of the compute scales considered, CPT from scratch performs much worse than pretraining on the full data, or split training.  Split training early at around 40B tokens performs similarly to pretraining over the full data.  Although we do not train a model that matches exactly the optimal number of tokens, we see that on either side of the optimal $t_s$, the accuracy is increasing then decreasing indicating that the optimal $t_s$ is a good indication of performance on language modeling and QA benchmarks. We report results for 125M models in Section~\ref{sec:125M_optimal} where we find better performance from split training with zero pretraining matching our results in Section~\ref{sec:synthetic} as the 125M models are sufficiently overtrained even at the earliest split training checkpoint.

\subsection{Optimal Model Split Point for Varying Sizes}
\textbf{Methodology:} We investigate split model training performance as a function of model size.  We repeat the analysis above, but for several model sizes from the GPT family of models.  Details of the architectures are in Appendix~\ref{sec:hyperparams}.  We compare pretrained models and split models at various scales throughout training.

\begin{figure}[t]
    \centering
    \begin{minipage}{0.55\textwidth}
      \centering
      \includegraphics[width=\textwidth]{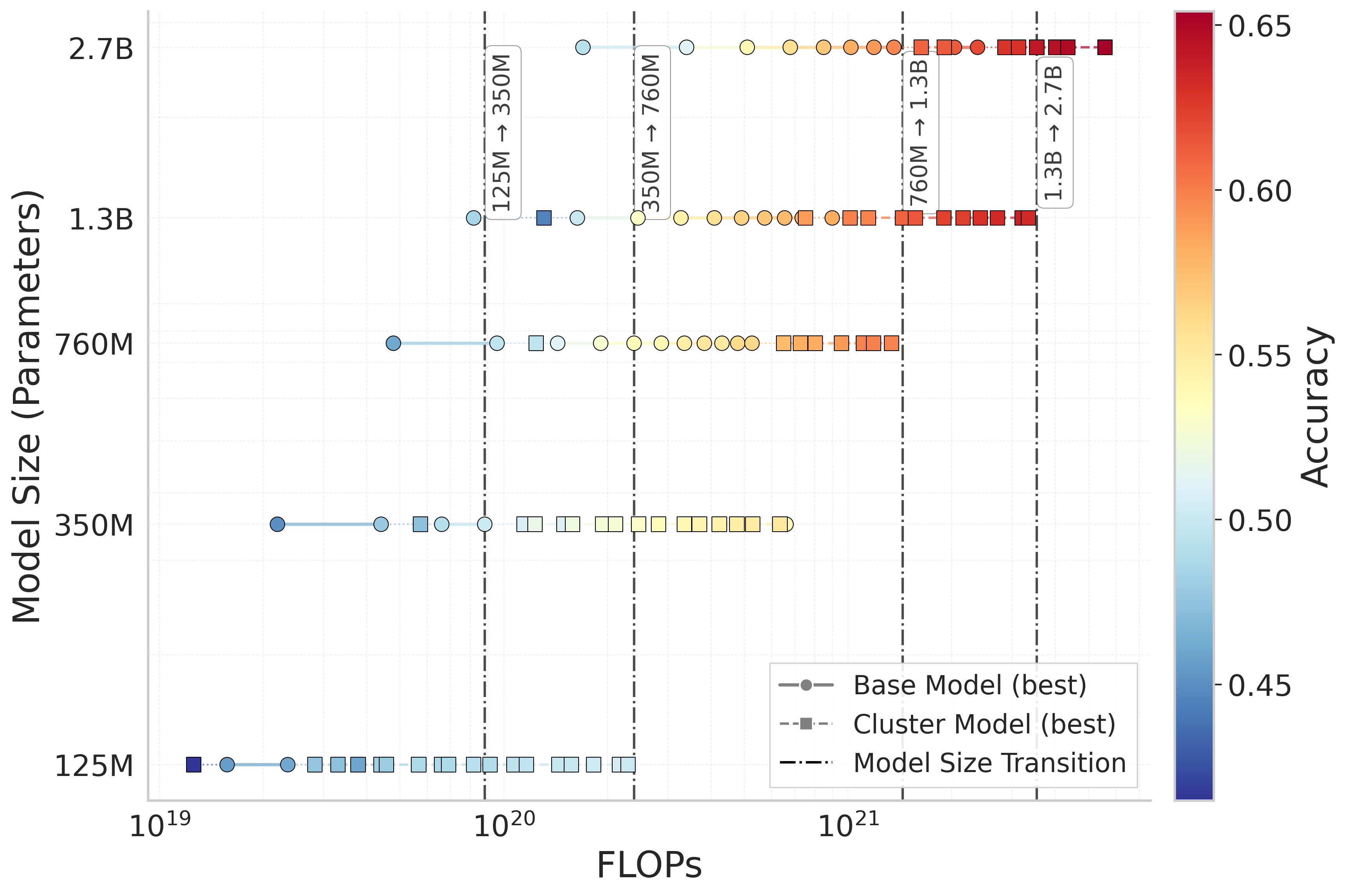}
      \subcaption{Optimal model size and splitting at varying flops.}
      \label{fig:optimal_ms_flops}
    \end{minipage}
    \hfill
    \begin{minipage}{0.40\textwidth}
      \centering
      \includegraphics[width=\textwidth]{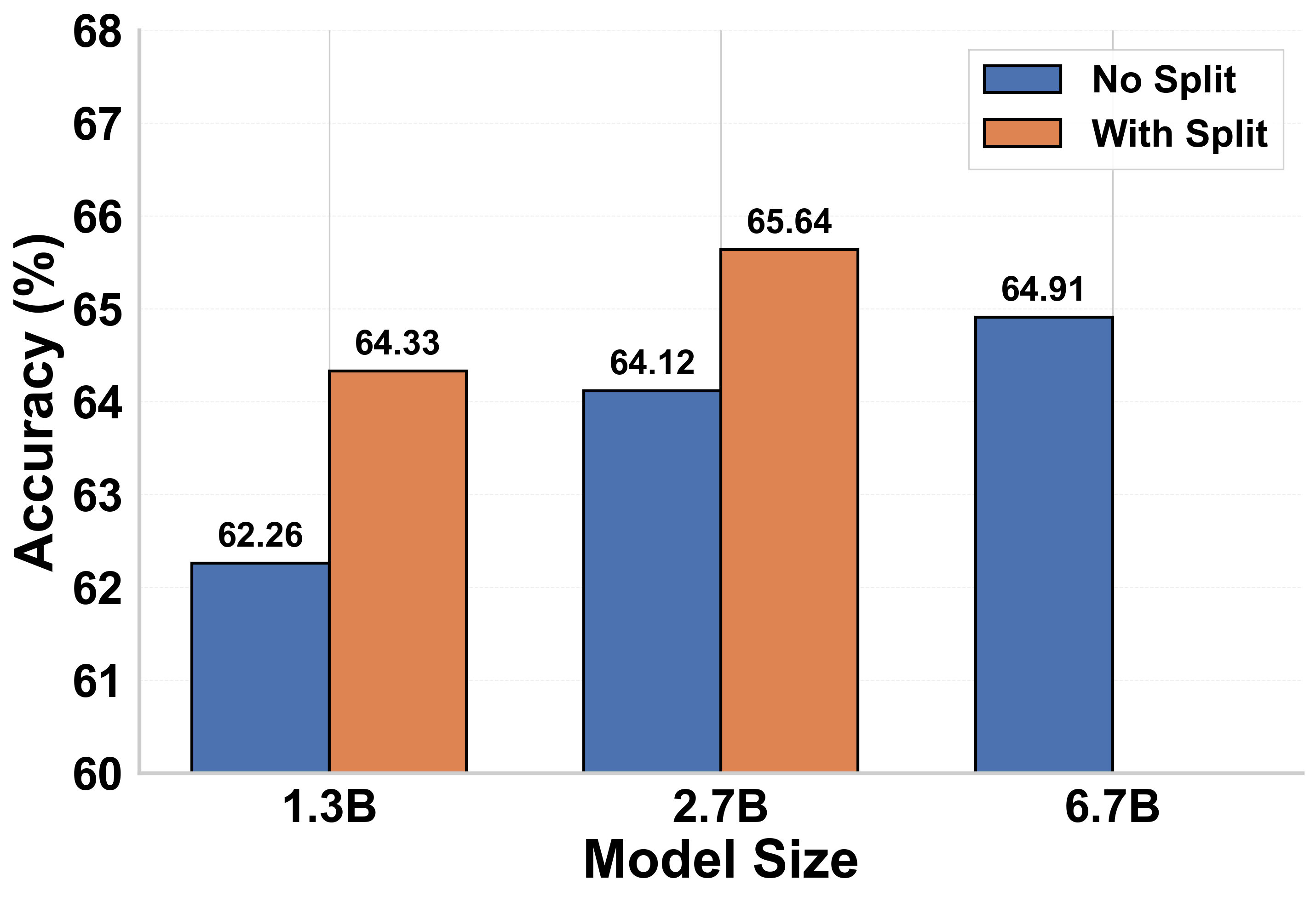}
      \subcaption{Average zero shot accuracy for pretrained and split models at the optimal allocation.}
      \label{fig:optimal_split_barplot}
    \end{minipage}
    \caption{Performance comparison across model sizes for split training. (a) Zero-shot QA accuracy for standalone (base) and split (cluster) models at various model sizes.  Dashed lines indicate where the best performing model changes in size as a function of the total compute (FlOPs).  (b) Zero-shot QA accuracy for various standalone and split training models at 1.3B and 2.7B scale.  All models are trained for a total compute budget equivalent to roughly 140B tokens of the 6.7B model, however 1.3B and 2.7B models have lower inference costs.}
    \label{fig:combined_optimal}
  \end{figure}



\textbf{Findings:} Results are summarized in \cref{fig:optimal_ms_flops}.  First, we observe that for a fixed model size (fixed line along the y-axis), the best performing model is a seed model (denoted by a circle), and changes to a split model (square) after which point split models always outperform the full pretrained model.  Second, we note that for several model sizes, such as 125M, 760M, and 1.3B, the split model can outperform the larger model size as the same amount of compute.  This is denoted by the switch point (vertical dashed lines) where the split models at the given size ar better than full pretrained models of the size higher. Following this observation, we note that in most cases, training a single pretrained model is rarely the best.  The only instance where the optimal model switches to a fully pretrained model at the next model size is from 350M to 760M at which point the best configuration is pretraining a single model.  At all other points of compute, split models are competitive or better.

\subsection{Optimal Pretraining Tokens Performance}
 We verify that using our scaling law estimates optimal $t_s$ in language modeling over specialized domains, and on zero-shot QA benchmarks.  
 
\subsubsection{Pile Language Modeling}
\textbf{Methodology:} We evaluate 1.3B models pretrained for the entire compute budget and split model training with $K=16$ domains from clustering the DCLM dataset.  We train for a compute budget of 720B tokens of the base pretrained model.  The split models are first pretrained for $t_s=340$B tokens before continued pretraining for the remainder.  We evaluate perplexity across five specialized domains in the Pile\footnote{\url{https://huggingface.co/datasets/monology/pile-uncopyrighted}}: ArXiv, DM Math, FreeLaw, Github, and PubMed Central \citep{gao2020pile}.  Details on the cluster distribution for the specialized domains are in Appendix~\ref{sec:pile}.   
 
\textbf{Findings:}  For split models, we compute the perplexity for each domain using the model routed to most frequently for data from that domain.   We compare with base pretraining only  in \cref{tab:ppl_pile} in Appendix~\ref{sec:pile}.  Split models perform better than the pretrained model with an average relative improvement in perplexity of 9.33\%.  The largest gains are from FreeLaw and ArXiv.  

\subsubsection{QA Benchmarks}
\textbf{Methodology:} We compare models using our scaling law to determine the optimal $t_s$ for training models, and train for a compute budget equivalent to $1\times$ Chinchilla for the 6.7B model equating to roughly 140B tokens as in \citep{li2024datacomp}.  We compare pretraining 1.3B and 2.7B parameter models, and split model training with the same size models.  Note that while the number of parameters is larger, the number of active parameters is much smaller than that of a 6.7B model. 

 

  \textbf{Findings:} We use our scaling law to estimate the optimal allocation $t_s$.  Results for 1.3B, 2.7B, and a pretrained 6.7B model are in \cref{fig:optimal_split_barplot}.  For the 1.3B parameter model this equated to 340B tokens for pretraining, less than half of the total pretraining tokens, and 220B tokens for the 2.7B parameter model.  At this scale, split models perform better than full pretraining by $1.5-2\%$.  The 1.3B split models perform worse than the 6.7B model, however, the 2.7B split models perform better by $0.6\%$.

\subsection{Ablations}
Prior sections split train models with a fixed number of domains $K=16$ and use all models at evaluation.  Naturally there are several important questions to consider.  In this section, we discuss whether all models are needed for inference, what the optimal number of domains is.

\subsubsection{Two Cluster Routing Specialization}
\textbf{Methodology:} Prior results train one model for each domain and route to all models at inference time.  In some settings it may be infeasible to maintain a model for each domain in memory.  In this section, we study specialization of domains at inference time where only two split models are used.  This reduces the number of total parameters greatly.  To select the domain models, we cluster the training sets of each of the QA tasks.  We take the top two clusters from each dataset for a total of seven unique clusters. We refer to this as Task Subset.  Alternatively, we could select a subset based on a single task which may be representative of a broader range of tasks.  To this end, we also consider an MMLU subset formed from the top two clusters from the MMLU train set.

\textbf{Findings:} Our findings in \cref{tab:subset_comparison} show that task specialization at evaluation reduces performance by around 1\% compared to evaluation and training with the full clusters. The MMLU subset reduces performance more by up to 3\%.  This indicates that performing well across domains necessitates all models, however some task specialization at inference time still improves over a base model in situations where the total number of parameters may be limited.  


  \begin{table}[t]
    \centering
      \centering
      \begin{tabular}{lccc}
      \toprule
       & \multicolumn{2}{c}{\textbf{Subset Routing}} & \textbf{Full} \\
      \cmidrule(lr){2-3} \cmidrule(lr){4-4}
      \textbf{PT Tokens} & MMLU Subset & Task Subset & Full \\
      \midrule
      10B  &  53.85  & 55.85   & 56.94   \\
      20B  & 55.68   & 57.05   & 58.11   \\
      40B  & 57.47   & 59.08   & 59.53   \\
      80B & 59.41 & 60.70 & 61.26 \\
      120B & 59.62 & 60.97 & 61.53\\
      \bottomrule
      \end{tabular}
    \caption{Routing comparisons for split training of a 1.3B model at 150B tokens.  Subset routing uses only two models per task at inference time whereas full routing uses all domain models.}
    \label{tab:subset_comparison}
  \end{table}




\subsubsection{Performance with Varying Number of Domains}
\textbf{Methodology:} Our main results use 16 domains, however data may come with any number of domains.  When the number of domains is small, the similarity may be large between the clusters, but the multiplicative factor to the number of steps will be reduced compared to more specific clusters. In contrast when the number of domains are large, each individual step on a domain improves the domain, but comes at a significant compute cost. We compare performance of split model training across  number of domains.  We train a separate clustering for each number of clusters. 

\begin{figure}
    \centering
    \includegraphics[width=\linewidth]{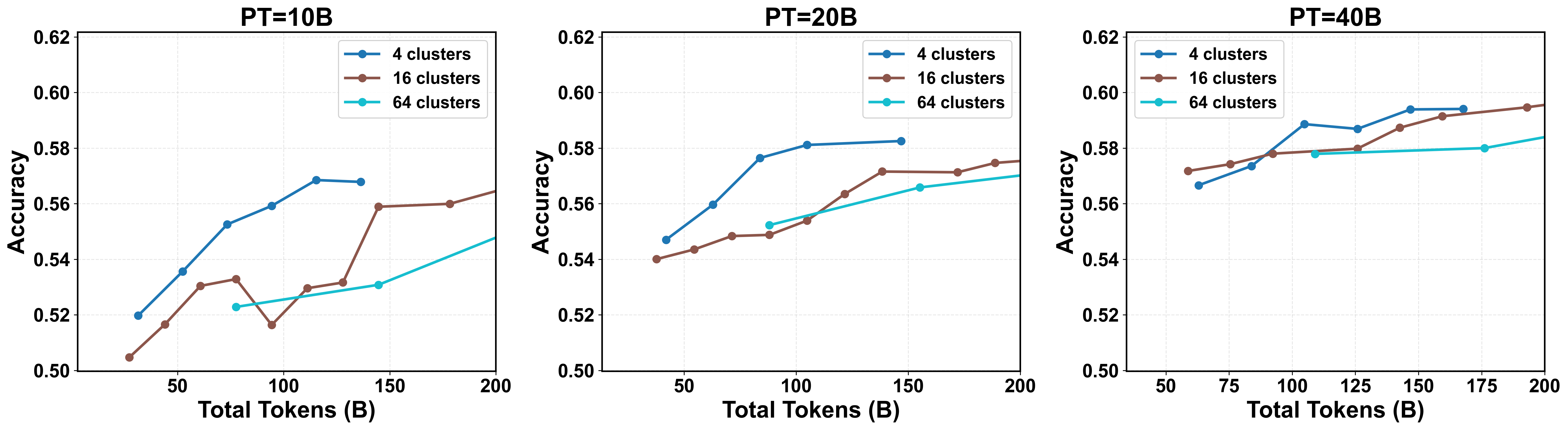}
    \caption{Accuracy vs. number of training tokens for varying number of domains and varying number of PT steps. All domain models are 1.3B parameter models.  The total number of parameters varies with the number of domains, but the number of active parameters remains constant at a single domain model.}
    \label{fig:1_3B_cluster_comparison}
\end{figure}

\textbf{Findings:} Results for training up to 200B tokens for the 1.3B parameter model are summarized in \cref{fig:1_3B_cluster_comparison}.  We find that the 16 domain model maintains a good balance in accuracy at small and large number of PT tokens.  At small number of PT tokens, the 16, and 64 domain models achieve low performance, and the 4 domain models perform the best. At larger $t_s$, the 4 domain models are matched by the 16 domain models.  Results for the 2.7B parameter model training up to 300B tokens are summarized in \cref{fig:2_7B_cluster_comparison}.

\section{Conclusion}
Strong language models are being trained in two stages through general pretraining on a large corpus of data before continuing on high quality specialized data for downstream applications. This work discusses split model training as a recipe for training language models when there are multiple specialized data domains in the pretraining corpus.  We propose a scaling law that directly addresses the question of how to allocate compute budget between pretraining on the entire dataset and continued pretraining on individual domains, and demonstrate that this is closely connected to downstream performance on zero-shot evaluation tasks.  

The gain of training independent expert models lies in both paradigms: (i) improved performance from reduced training on extraneous domains, and (ii) greater training efficiency from reduced hardware constraints.  Empirically, our scaling law accurately predicts performance of large models beyond the size trained and with additional tokens, and training language models with the optimal splitting point yields 2\% improved performance across different compute budgets and model sizes in evaluations leading to greater performance at the same compute budget as models with double the inference cost.  For the latter, specialized hardware (synchronous compute and large numbers of GPUs) are typically needed in order to train a single model, and there is a limit to the number of GPUs that can be used, especially for training smaller models due to limits in the batch size used for training.  Split model training for larger compute budgets requires less training on specialized hardware as in some settings less than 50\% of the compute budget is allocated to pretraining.  For the remainder of training,  models can be trained asynchronously and over less specialized hardware.

Additionally, there are other benefits to split model training over data domains beyond what we study here, for example in safe data handling.  In this setting, it may be undesirable to share data from particular domains, or have a model trained on a mix of data from different sources as data attribution becomes challenging.  Split training on a portion of shareable data during the PT stage and the private data during the CPT stage can make this feasible.  In summary, it is important for researchers and practitioners to consider how to allocate compute towards training the next large language models. In many instances, pretraining and continued pretraining are considered independently, however greater improvements come from considering them jointly.

\section*{Acknowledgements}
We are grateful to Zak Aldeneh, Natalie Schluter, Anastasiia Sedova, and Maartje ter Hoeve for their helpful discussions, comments, and thoughtful feedback in reviewing this work.

\bibliography{main}
\bibliographystyle{plainnat}

\clearpage
\appendix
\section{Training and Model Hyperparameters}
\label{sec:hyperparams}

The 100M parameter model consists of 24 layers, 8 attention heads, and a hidden dimension of 512. It is trained with a learning rate of 0.0003, weight decay of 0.01, and gradient clipping of 0.1. The 125M parameter model consists of 12 layers, 12 attention heads, and a hidden dimension of 768. It is trained with a learning rate of 0.001, weight decay of 0.0, and gradient clipping of 0.1.  The 350M  parameter model consisting of 24 layers, 16 attention heads, and a hidden dimension size of 1024. It is trained with a learning rate of 0.0003, weight decay of 0.01, and gradient clipping of 0.1. The 1.3B  parameter model consists of 24 layers, 16 attention heads, and a hidden dimension size of 2048.  It is trained with a learning rate of 0.0003, weight decay of 0.01, and gradient clipping of 0.1. The 2.7B parameter model consists of 32 layers with 2560 hidden dimension and 32 attention heads. It is trained with a learning rate of 0.00016, weight decay of 0.01, and gradient clipping of 1.0.  The 6.7B parameter model consists of 32 layers, 32 attention heads, and a hidden dimension fo 4096.   It is trained with a learning rate of 0.00012, weight decay of 0.01, and gradient clipping of 1.0.  

For our results in Section~\ref{sec:scaling_law_fitting} and \ref{sec:exp_lm}, all pretrained-only models are trained with a warmup of 10000 steps with fixed learning rate.  Split models are trained with a fixed learning rate starting from the learning rate of the base model with the optimizer states reset to mirror CPT from a given pretrained model.  If splitting from the start of training, we use a warmup of 3000 steps.  For models trained in Section~\ref{sec:synthetic}, we use a warmup set to 1\% of the training time.   All models are trained with the AdamW optimizer using $\beta_1=0.9$ and $\beta_2=0.999$.  Unless otherwise stated, models use the GPT2 tokenizer, with a total vocabulary size of 50K tokens using BPE.

All dense models are trained using NVIDIA’s Megatron-LM\footnote{\url{https://github.com/NVIDIA/Megatron-LM}} repository for pretraining language models. 

For the MoE models, we train with AdamW (weight decay $0.1$) using a learning-rate schedule with $5\%$ linear warm-up followed by decay over the final $10\%$ of training, for a total of $260$B tokens. To obtain compute-matched clustered runs as in experiments with dense models, we start from the baseline checkpoint at step 150{,}000 and train $K=16$ cluster models. Each cluster model is then continued for 6{,}250 steps so that the aggregate continuation across all clusters matches the baseline continuation from 150{,}000 to 250{,}000 steps (and thus matches total tokens and training FLOPs under the same per-step batch size and sequence length).  We use the megablocks framework for training \citep{gale2023megablocks}.

\section{Training Cost}
A majority of the costs for running our experiments are for 1.3B and 2.7B models.  The total training time for 1.3B models on roughly 100B tokens is around 1000 GPUh on Nvidia H100 GPUs.  For a 2.7B model trained on 100B tokens, the total time is around 2200 GPUh on Nvidia H100.

For experiments in \cref{fig:optimal_split_barplot}, this would equate to 7000+ GPUh for each model.  For a model which is fully pretrained, this would involve training on 42 GPUs for a full week, which can be prohibitively expensive for many practitioners or researchers training language models.  Training split models instead means that 47\% of the total budget will be spent on pretraining.  The remaining time can be spent training independently on non-specialized hardware as each split model individually needs less compute.  

\section{Datasets}
\subsection{Training Datasets}
We pretrain all models using the DCLM-Baseline dataset \citep{li2024datacomp}.  We train on a random shuffled subset of  around 400B tokens of data, which are split into up to 2049 sized chunks for clustering following \cite{grangier2025task,pouransari2025pretraining}. Except for \cref{fig:optimal_split_barplot} where the 1.3B parameter model is trained on $<2$ full repetitions of the data.  At the 1.3B model scale, we do not suspect this leads to a decrease in performance.  For cluster models, we cluster the full corpus, but typically only train on a small portion of the data not exceeding $100$B tokens in any of our experiments.

\subsection{Evaluation Datasets}
\label{sec:evaluation_datasets}
\subsubsection{Validation Loss}
We hold out a set of 10,000 documents from each cluster and evaluate the trained models on each document.  We ensure that these documents are also not seen in the pretraining data.  We use the validation set from Pile uncopyrighted.  

\subsubsection{Zero Shot Evaluations}
\begin{itemize}
    \item \textbf{SciQ}: A dataset of science exam questions for evaluating the ability of NLP models in understanding and reasoning within the science domain~\citep{welbl2017crowdsourcing}.
    \item \textbf{ARC Challenge (ARC-C)}:Part of the AI2 Reasoning Challenge (ARC)~\citep{clark2018think}, containing science exam questions from grades 3 to 9. The ARC Challenge set includes more difficult questions that necessitate higher-order reasoning.
    \item \textbf{ARC Easy (ARC-E)}: The Easy set of the AI2 Reasoning Challenge~\citep{clark2018think} features questions from the same source as ARC-C but are considered less challenging.
    \item \textbf{Winogrande (WG)}: This dataset challenges models on common sense reasoning in a language context, focusing on pronoun disambiguation tasks~\citep{sakaguchi2021winogrande}.
    \item \textbf{PIQA}: Physical Interaction Question Answering tests the understanding of everyday physical processes~\citep{bisk2020piqa}.
    \item \textbf{HellaSwag (HS)}: Evaluates a model's ability to complete scenarios in a contextually and logically coherent manner~\citep{zellers2019hellaswag}.
    \item \textbf{BoolQ}: A set of Yes/No questions from Google Search Queries. Samples contain a passage, and question \citep{clark2019boolq}.
    \item \textbf{MMLU}: Multi-domain question answering, MMLU assesses the model's expertise over a wide range of specialized subjects, from professional domains to academia~\citep{hendrycks2020measuring}.
\end{itemize}

\section{Evaluation Details}
We use the lm-eval-harness repository\footnote{\url{https://github.com/EleutherAI/lm-evaluation-harness}} for zero-shot accuracy on QA tasks\footnote{We use the commit \texttt{03c44adc0586f88bb343a74da1a1c602103536dd}.}.  All datasets are collected from the Huggingface datasets library.  For all tasks, we use the continuation (cloze) style formatting for evaluation.

For split model selection over the evaluation set, we route using either the question, or passage.  \cref{tab:routing-fields} summarizes the routing query for downstream evaluation tasks.  For validation set evaluaiton, we evaluate using the first 32 tokens of the document, which would correspond to around 24 words using an estimate of 0.75 words per token.  

\begin{table}[h]
  \centering
  
  \begin{tabular}{ll}
  \toprule
  \textbf{Task} & \textbf{Routing Text} \\
  \midrule
  ARC & question \\
  SciQ & question \\
  HellaSwag & ctx \\
  PIQA & goal \\
  BoolQ & question \\
  Winogrande & sentence \\
  MMLU & question \\
  \bottomrule
  \end{tabular}
  \caption{Routing Text Field by Task}
  \label{tab:routing-fields}
  \end{table}

\section{Cluster Details}
\label{sec:cluster_details}

\subsection{Cluster Implementation Details}
We cluster the training set using a balanced K-means clustering algorithm  with a tree of depth 1 as we train with a maximum of 64 clusters following \citep{grangier2025task}. Before training the clustering, the dataset is segmented into non-overlapping 2,048 token windows and compute sentence-BERT embedding for every windows following \citep{pouransari2025pretraining}. Embedding are built from the sentence-BERT MiniLM-L6-v2 model \citep{reimers-2019-sentence-bert}.

\subsection{Pile Distribution and Loss Over Clusters}
\label{sec:pile}
\begin{figure*}[htbp]
      \centering
      \begin{subfigure}[b]{0.48\textwidth}
          \centering
          \includegraphics[width=\textwidth]{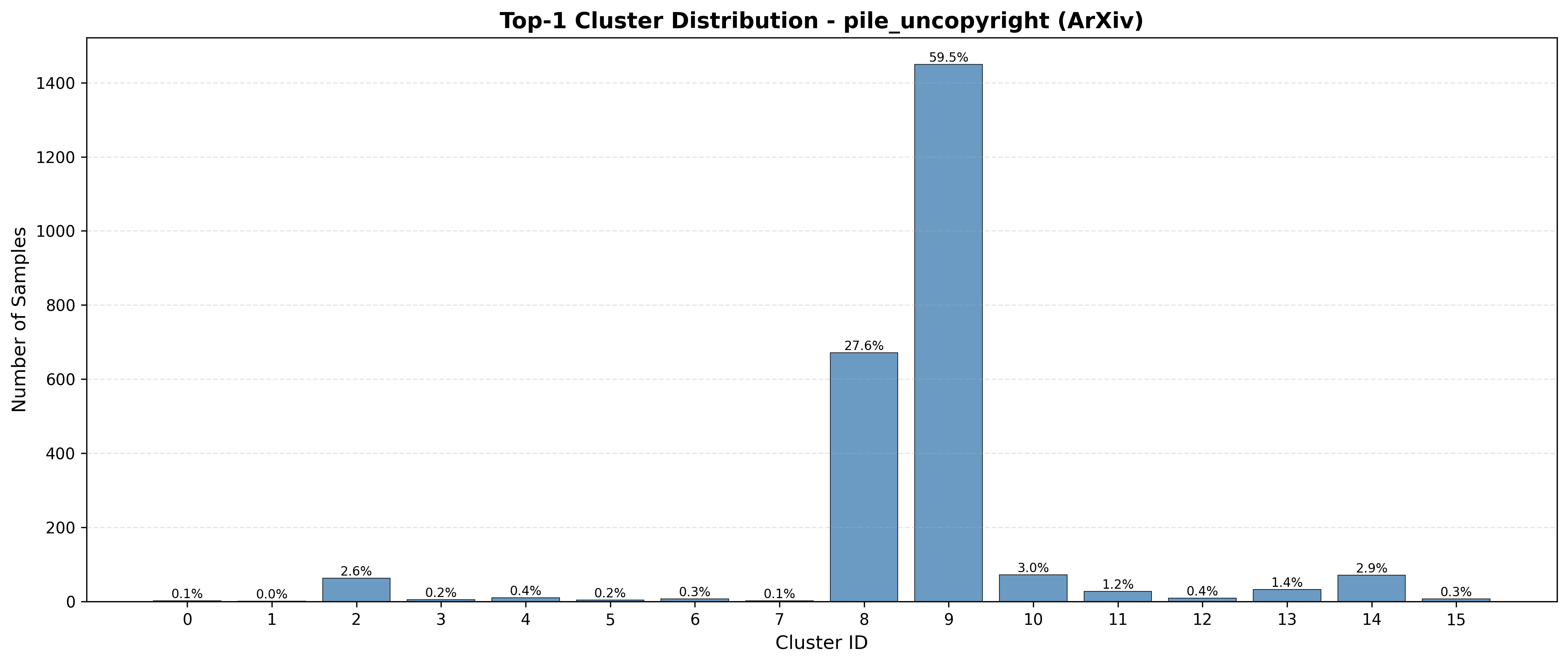}
          \caption{ArXiv}
          \label{fig:appendix-arxiv}
      \end{subfigure}
      \hfill
      \begin{subfigure}[b]{0.48\textwidth}
          \centering
          \includegraphics[width=\textwidth]{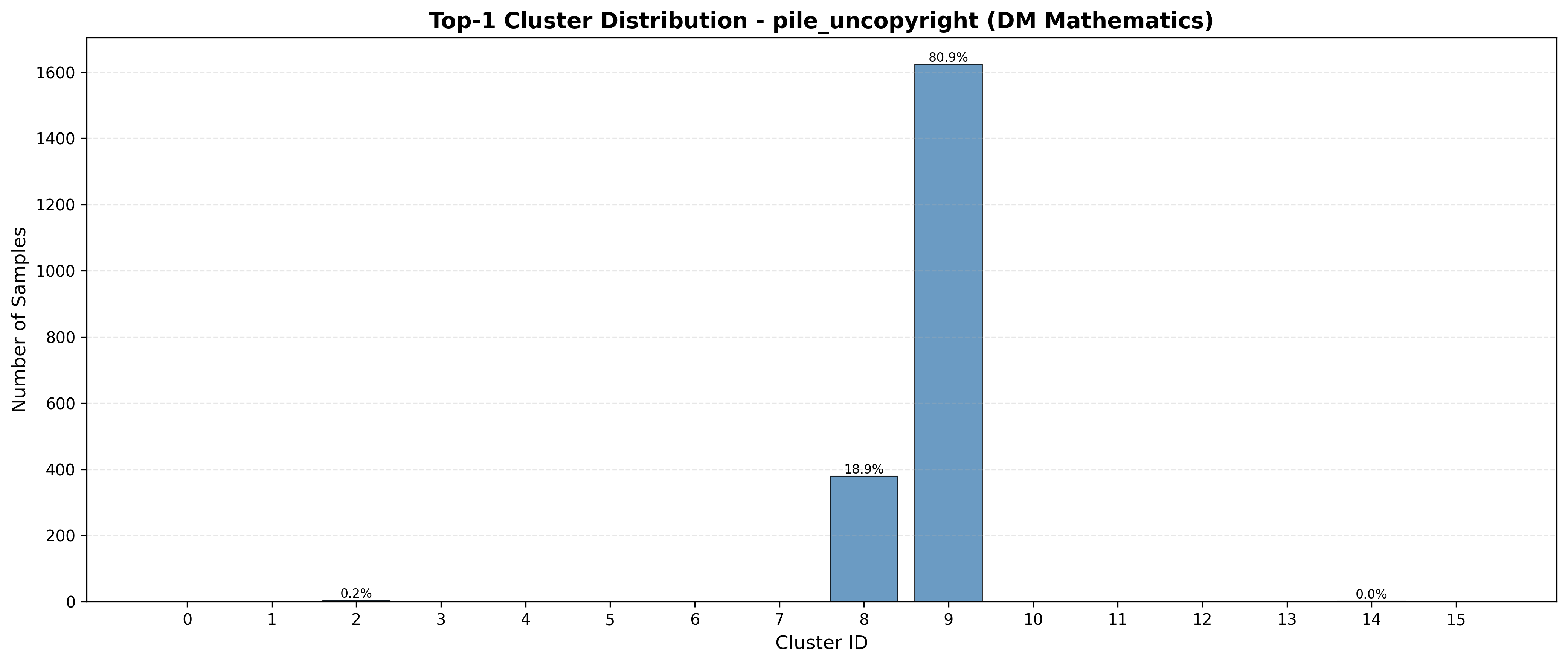}
          \caption{DM Mathematics}
          \label{fig:appendix-math}
      \end{subfigure}

      \begin{subfigure}[b]{0.48\textwidth}
          \centering
          \includegraphics[width=\textwidth]{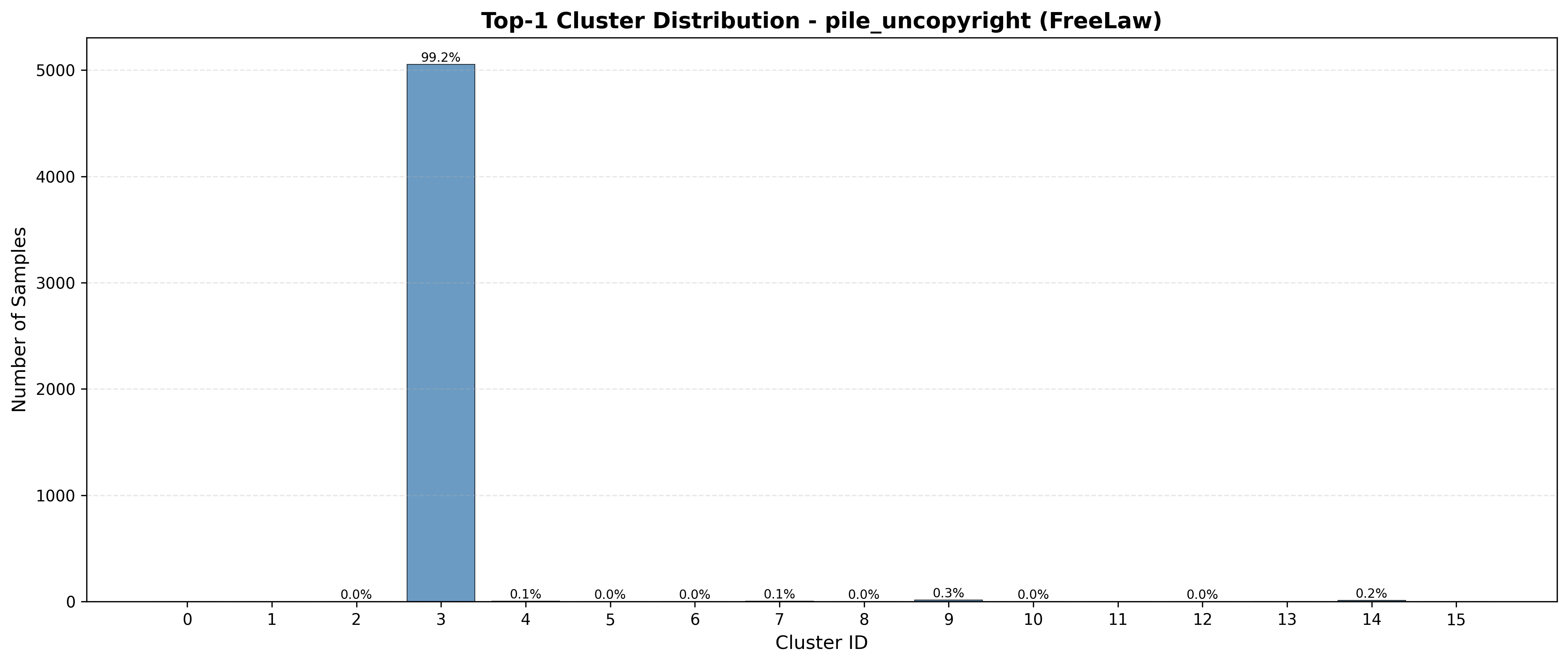}
          \caption{FreeLaw}
          \label{fig:appendix-freelaw}
      \end{subfigure}
      \hfill
      \begin{subfigure}[b]{0.48\textwidth}
          \centering
          \includegraphics[width=\textwidth]{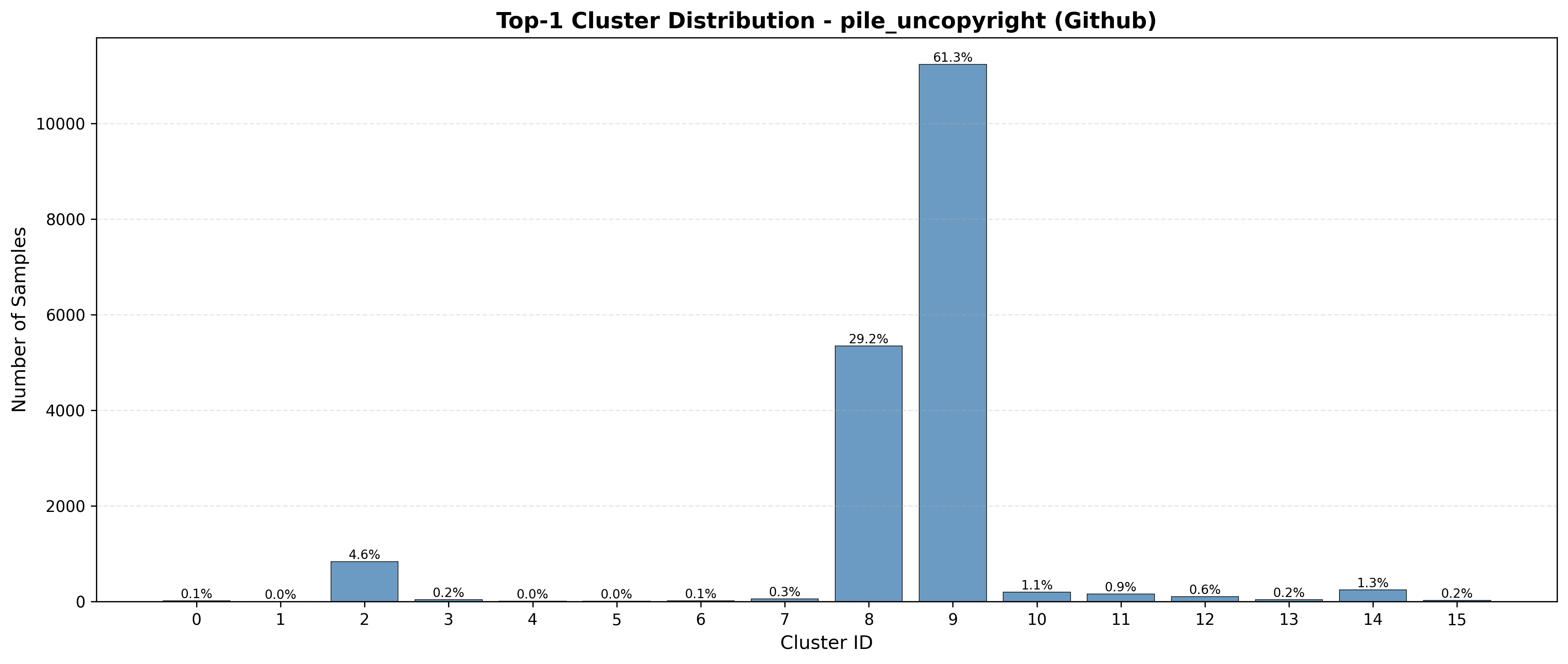}
          \caption{Github}
          \label{fig:appendix-github}
      \end{subfigure}

      \begin{subfigure}[b]{0.48\textwidth}
          \centering
          \includegraphics[width=\textwidth]{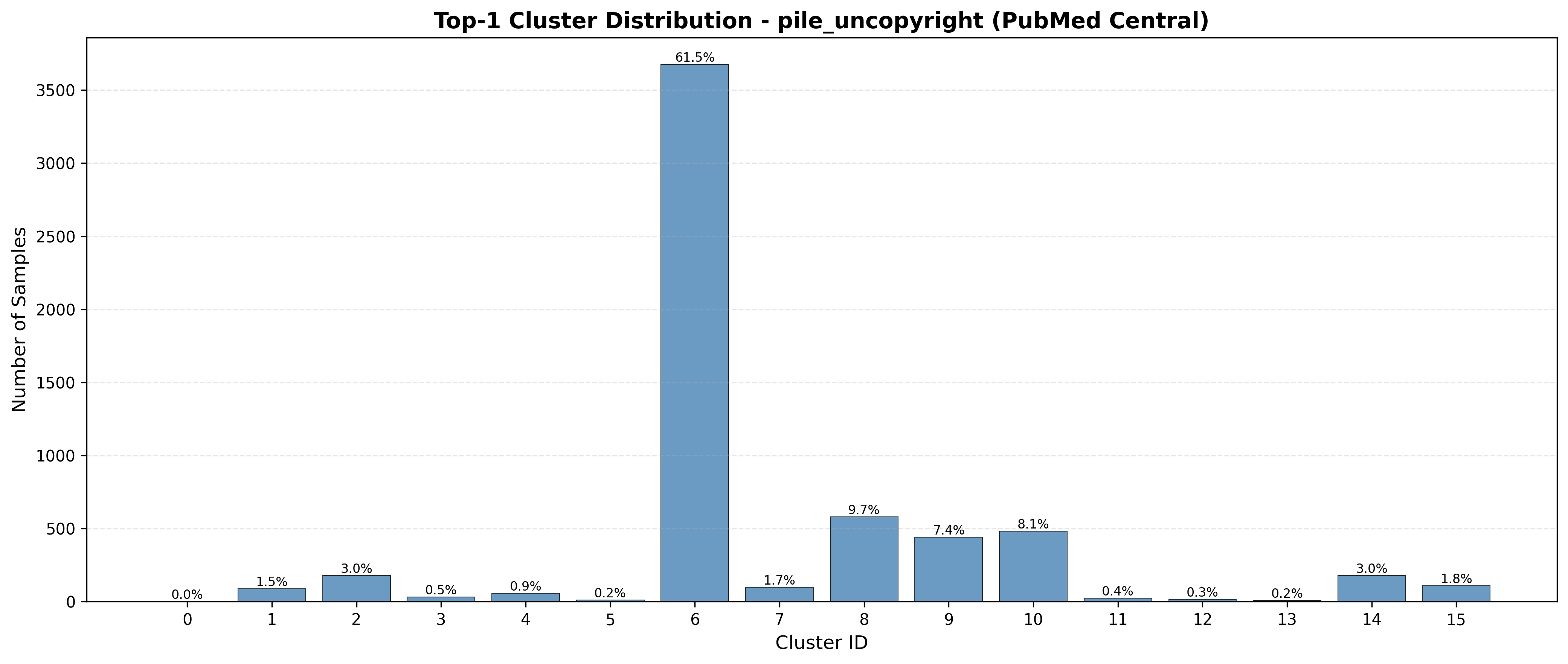}
          \caption{PubMed Central}
          \label{fig:appendix-pubmed}
      \end{subfigure}

      \caption{Distribution of samples from Pile domains within the DCLM clustering.}
      \label{fig:pile_cluster_dist}
  \end{figure*}

We first verify that the clustering produces specialized datasets and as a result specialized split models.  To do so, we measure distribution of samples from different domains of the Pile.  We consider ArXiv, DM Math, FreeLaw, Github, and PubMed Central for their diversity and relevance towards in many CPT settings such as training code, math, and medical language models.  The distribution is shown in \cref{fig:pile_cluster_dist}.

\begin{figure}
      \centering
      \includegraphics[width=0.75\linewidth]{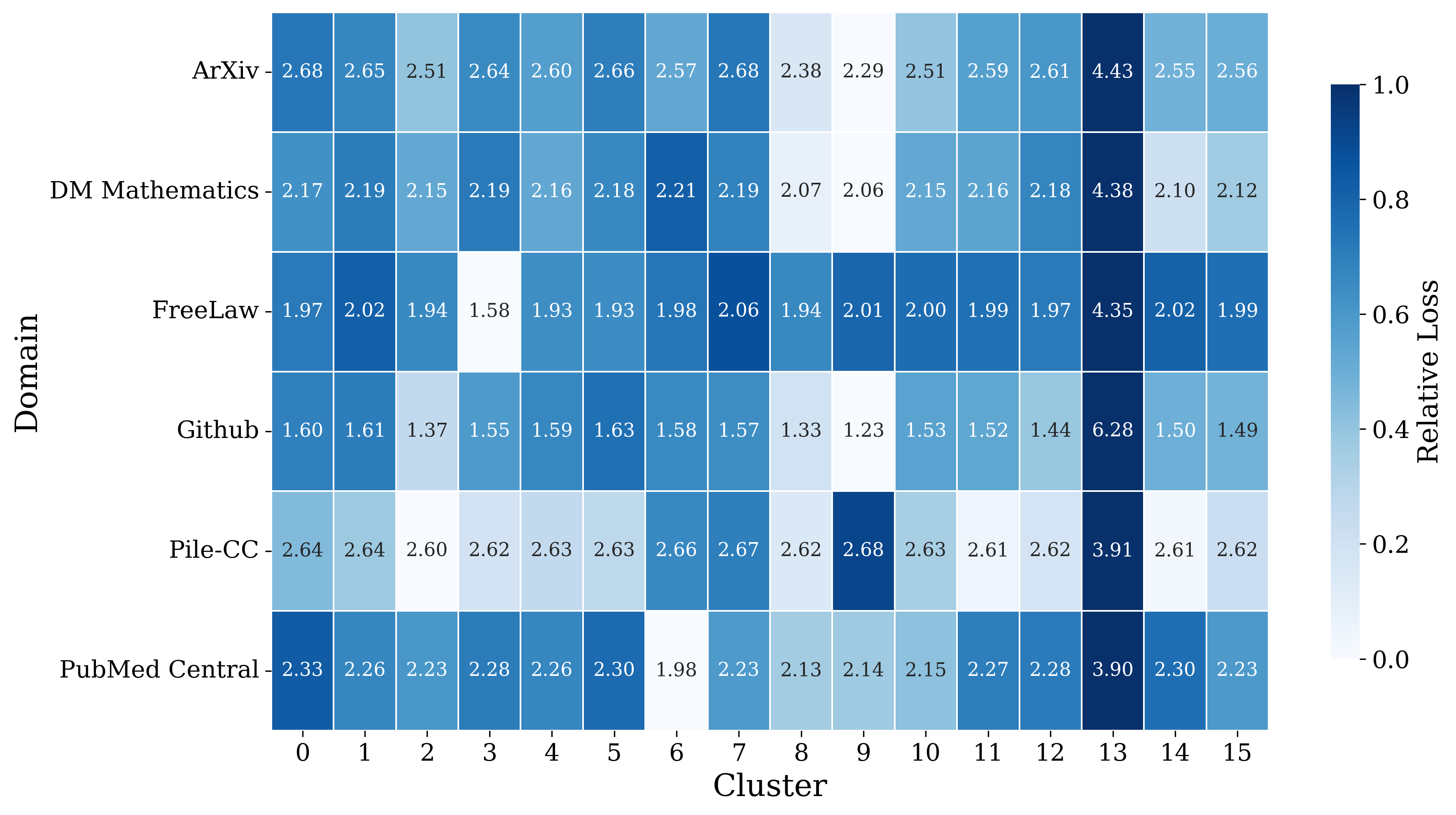}
      \caption{Loss for five domains of the pile and common crawl data for each split model.  The models are 1.3B models pretrained for 340B tokens before CPT.}
      \label{fig:pile_loss}
  \end{figure}

  Next, we compute the loss for each split model over the five domains in \cref{fig:pile_loss}.  To compute the loss we take a subset of 1000 samples from the validation set of each domain.

  Finally, we compare the perplexity of a base pretrained-only model trained for 720B tokens, and split models first pretrained for $t_s=340$B tokens.  We compute perplexity as 
  $$
  PPL(X) = exp\left( - \frac{1}{L} \sum_{t=1}^Lp(x_t | x_{<t}) \right)
  $$
  We route all sequences from a domain to a single model.  For some domains, better performance could be achieved as different samples are mapped to different clusters.  Nonetheless, even routing samples only to the most common cluster, we still see relative improvements across these five domains.  

\begin{table}
    \centering
    \begin{tabular}{l|c|c}
    \hline
     Domain &  Pretrained  & Split Model \\
     \hline
       ArXiv  & 11.01 & 9.92\\
       DM Math & 8.14 & 7.81\\
       FreeLaw & 5.76 & 4.85\\
       Github & 3.74 & 3.42\\
       PubMed Central & 7.90 & 7.25\\
       \hline
    \end{tabular}
    \caption{PPL for different domains of the Pile for a standalone pretrainged model only, and split models.  The standalone model with pretraining only is trained for a total of 720B tokens.  The split models are 1.3B parameters and trained for the same total compute budget with $t_s=340$B.}
    \label{tab:ppl_pile}
\end{table}

\subsection{Prefix Routing Accuracy}

At inference time in settings where the domain is unknown, routing is done using a prefix.  For QA tasks, this can be the entire query and context, or a portion of the query such as the question only.  For general language modeling, a small portion at the start of the document is used.  The loss is then computed over the remaining portion of the document.  In \cref{tab:prefix_retrieval}, we evaluate sensitivity of the clustering to different prefix lengths by measuring retrieval accuracy of the full document given the prefix.  Our findings show that 
retrieval accuracy is high starting at 16 tokens and diminishes in gain of R@5 starting from 32 tokens.  We choose 32 tokens as it is a good tradeoff and should be representative of the context length of downstream tasks equating to roughly 1-2 sentences.  
\begin{table}[h]
\centering
\begin{tabular}{ccc}
\toprule
Prefix Length (tokens) & R@1 & R@5 \\
\midrule
1 & 0.0128 & 0.0362 \\
2 & 0.0955 & 0.1753 \\
4 & 0.2664 & 0.3765 \\
8 & 0.4802 & 0.5884 \\
16 & 0.7112 & 0.8078 \\
32 & 0.8775 & 0.9330 \\
64 & 0.9663 & 0.9870 \\
128 & 0.9972 & 0.9995 \\
256 & 0.9998 & 1.0000 \\
\bottomrule
\end{tabular}
\caption{Retrieval accuracy by prefix length (N=10000)}
\label{tab:prefix_retrieval}
\end{table}

\section{Two Cluster DCLM Experiments}
\label{sec:two_cluster}
\textbf{Methodology:} We train language models with pairs of clusters from the clustered DCLM dataset.  We create three pairs of clusters by computing the distance of the centroids from all clusters to cluster 0.  We then take the two farthest clusters (denoted 0\% similarity), the median cluster distance (denoted 50\% similarity), and the two closest clusters (denoted 100\% similarity).  We compare split training on both clusters and training a single language model on both clusters combined.  

\begin{figure}
    \centering
    \includegraphics[width=\linewidth]{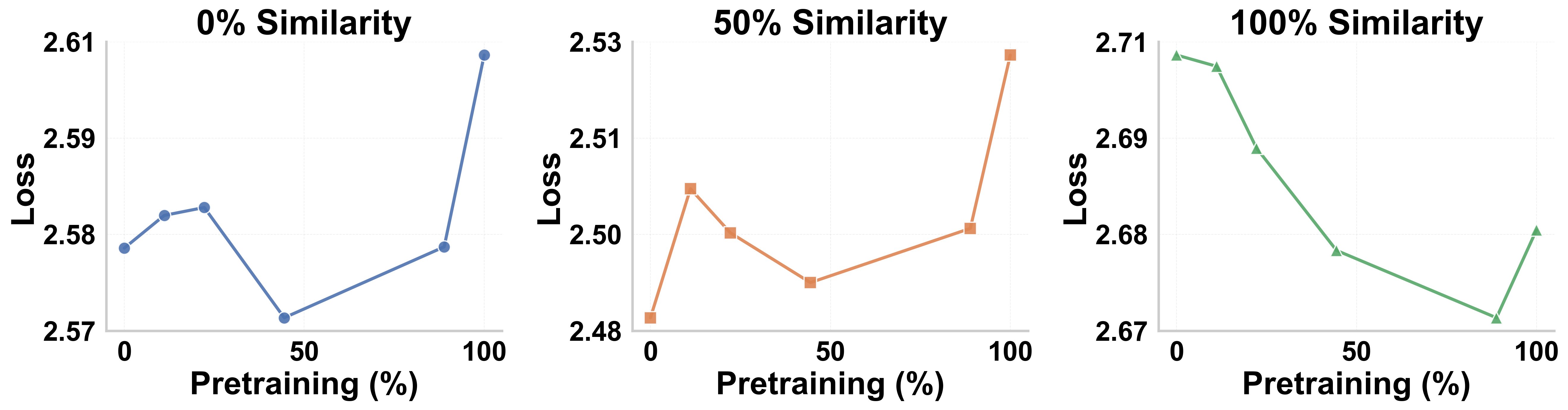}
    \caption{Average loss vs. fraction of compute for pretraining for different pairs of clusters in the DCLM-base dataset. Total compute budget is 90000 steps or roughly 90B tokens.}
    \label{fig:dclm_two_cluster}
\end{figure}

\textbf{Findings: } Results are shown in \cref{fig:dclm_two_cluster} for training at 90K steps (roughly $3\times$ Chinchilla). For the 0\% and 50\% similarity, the loss is mostly increasing with proportion of budget for pretraining.  For the 100\% similarity setting, we find the reverse where the loss decreases as the proportion of pretraining increases.  This matches the findings from memorization of phonebooks in the previous section.


\section{Scaling Law Estimation}

\subsection{Scaling Law Optimization}
\label{app:scalefit}
We fit on data from for base model training for  5B-200B tokens for the pretrained model, and between 5B-30B tokens for continued pretraining depending on the amount of training done in pretraining. We remove outliers from the data ($>4.0$ and $<0.5$).  We do not remove any steps and predict $0$ steps for either pretraining or continued pretraining.

We fit the scaling laws using the same method as in ~\citep{shukor2025scaling}. 
Given a sequence of parameters $N^i, D^i, D_k^i$ as well as the corresponding loss values $L^i$, for $i=1\dots n$, we find the optimal scaling law parameters $\Omega$ by solving 
$$
\min_{\Omega} \sum_{i=1}^n \mathrm{Huber}(\mathcal{L}(N^i, D^i, D_k^i; \Omega) - L^i)
$$
where $\mathrm{Huber}$ is the Huber loss with a shape parameter of $\delta =10^{-3}$. In order to solve that problem, we resort to the Basin-Hopping algorithm with a grid of random initializations, which we found more efficient that the widely used L-BFGS approach.

Several parameters in the scaling law share overlapping functional roles, which can lead to identifiability issues during optimization. For example, $E_p$ and $E_0$ jointly determine the asymptotic loss, while $\alpha_1$, $\alpha_2$, and $c$ both modulate the effective contribution of pretraining tokens. Without constraints, the optimizer can find similar fits with very different parameter values. To regularize the optimization and ensure similar fits across clusters, data, and model sizes, we  bound each parameter to a plausible range. Specifically, we use softplus  for parameters that must be strictly positive, a scaled sigmoid for parameters confined to a known interval $[l, u]$, and the standard sigmoid for powers.  The parameter bounds are summarized in \cref{tab:param_constraints}.

  \begin{table}[t]
  \centering
    \begin{tabular}{l|l|l}
  \toprule
  Parameter & Transform & Effective range \\
  \midrule
  $E_p$ &  softplus & $(0, \infty)$ \\
  $E_0$  & scaled sigmoid & $[1.0,\; 3.0]$ \\
  $A$ & softplus & $(0, \infty)$ \\
  $\alpha_1$ & sigmoid & $[0.1,\; 1.0]$ \\
  $\alpha_2$ & sigmoid & $[0.1,\; 1.0]$ \\
  $c$ & sigmoid & $[0.5,\; 4.0]$ \\
  $D_s$ &  scaled sigmoid & $[500,\; 700]$ \\
  $\gamma$ & sigmoid & $[0.1,\; 1.0]$ \\
  \bottomrule
  \end{tabular}
  \caption{Parameter constraints for the single-size CPT scaling law. Transformations are applied
  to unconstrained optimizer variables to enforce bounded, physically meaningful parameter values.}
  \label{tab:param_constraints}
\end{table}

\subsection{Optimal PT Extrapolation}

\begin{figure}
    \centering
    \includegraphics[width=0.75\linewidth]{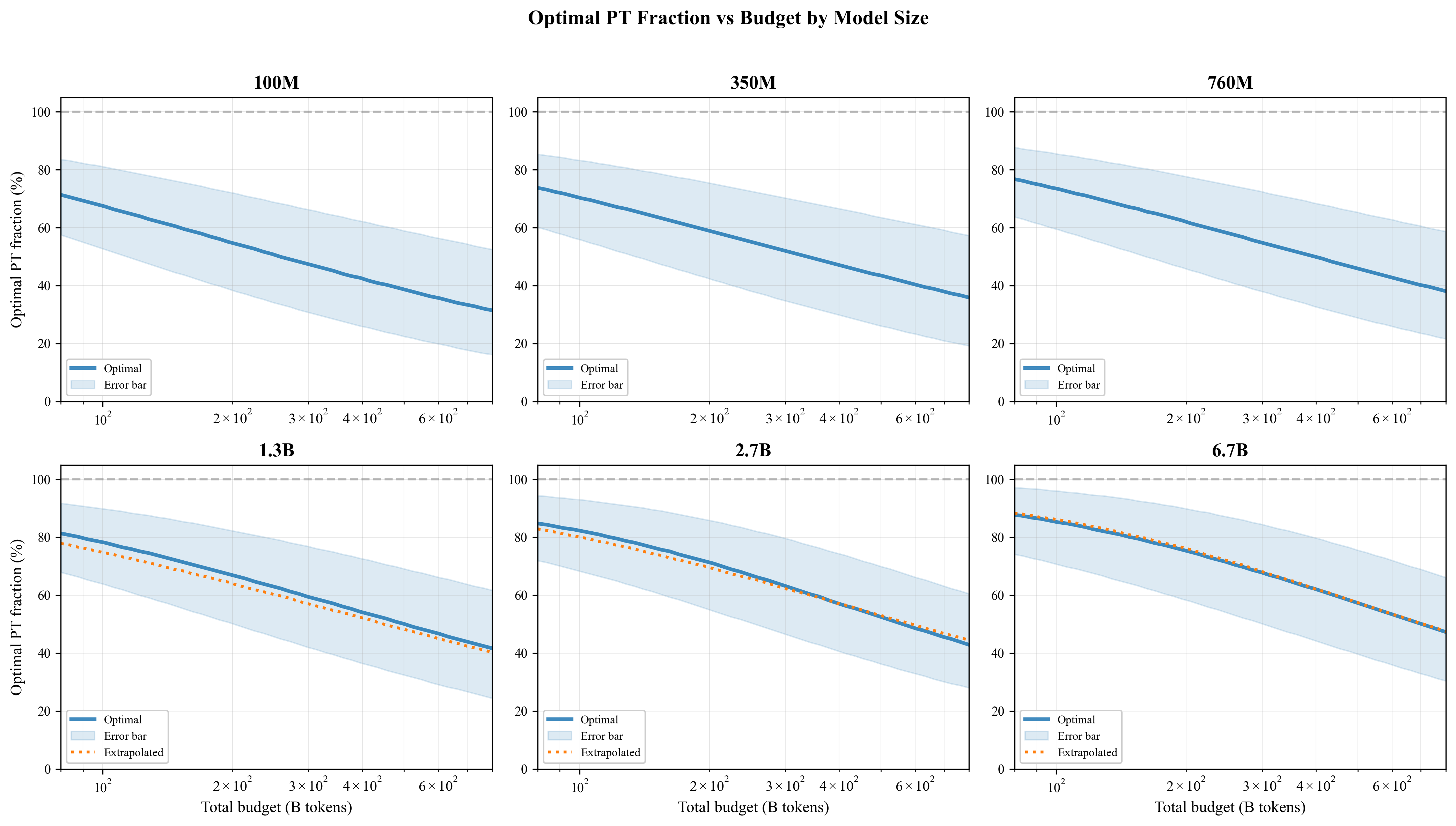}
    \caption{Estimated optimal fraction of pretraining tokens as a function of total compute budget, estimated from per-size scaling laws (solid lines) and extrapolation via power-law fit across smaller sizes (dotted lines). Shaded intervals represent the range of allocations that achieve loss within 0.005 of the optimum, capturing the flatness of the loss surface near the optimal split.}
    \label{fig:extrapolate_optimal_pt}
\end{figure}

We additionally present a simple extrapolation procedure that leverages scaling law fits from smaller models. For each model size $N_i$, we first fit a single-size scaling law $\hat{L}(D_{\text{pt}}, D_{\text{cpt}}; N_i)$ on observed loss data. Given a total compute budget $B$ expressed in effective tokens, where the budget constraint is $D_{\text{pt}} + K \cdot D_{\text{cpt}} = D_T$ (with $K=16$ reflecting the higher per-token cost of continued pretraining), we sweep over feasible allocations and identify the optimal pretraining fraction $f_s = t_s / D_T$ that minimizes predicted loss. For any fixed budget $B$, the optimal pretraining fraction varies smoothly as a function of model size (\cref{fig:extrapolate_optimal_pt}), motivating a power-law fit $f_s(N) = a \cdot N^{b}$ across model sizes.

To extrapolate to a target model size, we fit this power law using data from all smaller model sizes in the dataset. For example, using scaling laws fit on 100M, 350M, 760M, and 1.3B models, we extrapolate the optimal pretraining fraction to 2.7B. We compare the extrapolated prediction with the ground-truth optimal $f_s$ obtained by fitting a scaling law directly on data from the target model size in \cref{fig:extrapolate_optimal_pt}. The extrapolation closely matches the per-size ground truth, suggesting that practitioners can estimate the optimal PT-CPT allocation for a larger model from a small number of cheaper, smaller-scale experiments, without requiring a joint scaling law that spans all model sizes.

\section{Comparison with other scaling laws for continued pre-training}
\label{app:sec:comparison_liew}
\begin{figure*}[ht]
    \centering
\includegraphics[width=0.32\textwidth]{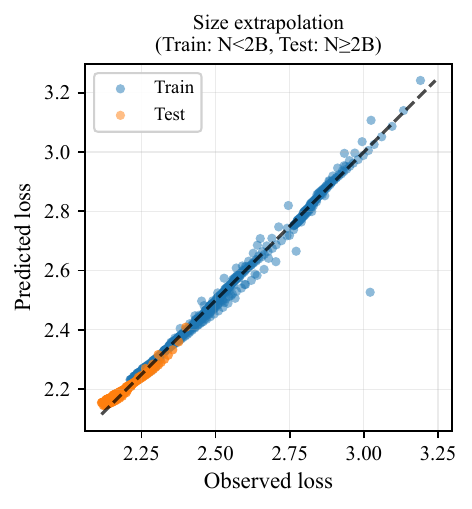}
\includegraphics[width=0.32\textwidth]{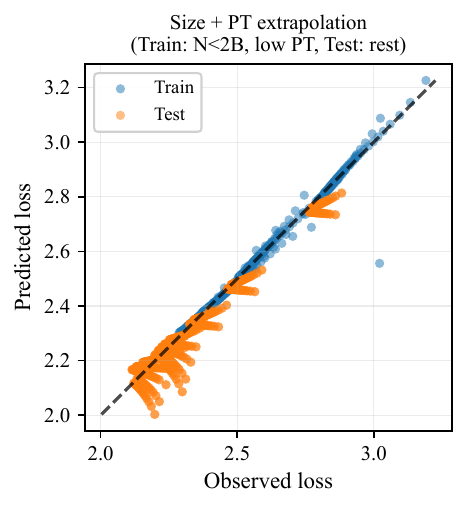}
\includegraphics[width=0.32\textwidth]{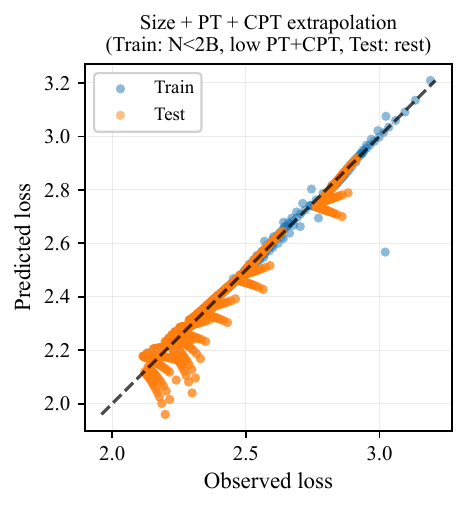}
    \caption{Estimated fits with the scaling law from \citep{liew2025acceleration}}
\label{fig:scaling_law_liew_plots}
\end{figure*}
\citet{liew2025acceleration} also tackle the problem of forecasting the loss of a model of size $N$, pretrained with $D$ tokens and then trained with $D'$ more token.
They propose a scaling law of the form
\begin{equation}
\label{eq:liew_scaling_law}
    L = E + A {D'}^{-\alpha_1}D^{-\alpha_2 +\alpha_3 \log D'} + B N^{-\beta}.
\end{equation}

We fit that scaling law to our data using the procedure described above.
We report the comparison of extrapolation capabilities in \cref{tab:liew_comparison}, and we observe that our law has a better predictive power.
We report the predicted vs actual loss plots in \cref{fig:scaling_law_liew_plots}.
We also note that the scaling law in \cref{eq:liew_scaling_law} does not verify the desiderata mentioned in Section~\ref{sec:functional_form}.

\begin{table}[ht]
    \centering
    \begin{tabular}{l|c|c|c|c}
    \hline
         & Train Size & Test Size & R2 (ours) & R2 \citep{liew2025acceleration} \\
       \hline
        Scenario 1 & 649 & 142 & \textbf{0.934} & 0.910 \\
        Scenario 2 & 344 & 447 & \textbf{0.982} & 0.946\\
        Scenario 3 & 156 & 640 & \textbf{0.979} & 0.953 \\
        \hline
    \end{tabular}
    \caption{Comparison of the test R2 scores with our proposed scaling law and that of \citep{liew2025acceleration}.}
    \label{tab:liew_comparison}
\end{table}

\section{Scaling Law Results for Other Clusters}

\subsection{Extended Results for Domain 0}

\begin{figure*}[ht]
    \centering
\includegraphics[width=0.32\textwidth]{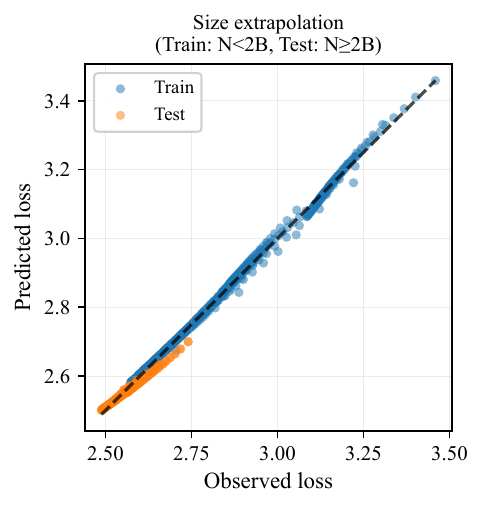}
\includegraphics[width=0.32\textwidth]{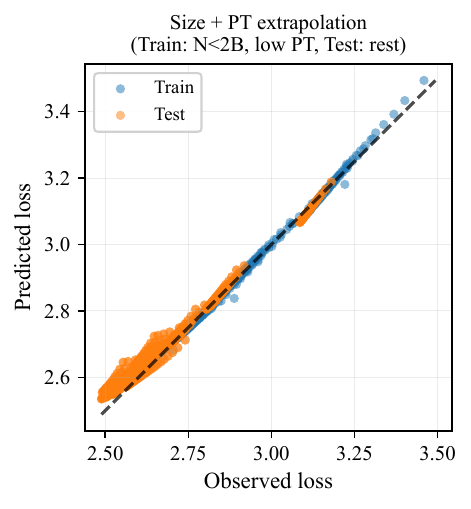}
\includegraphics[width=0.32\textwidth]{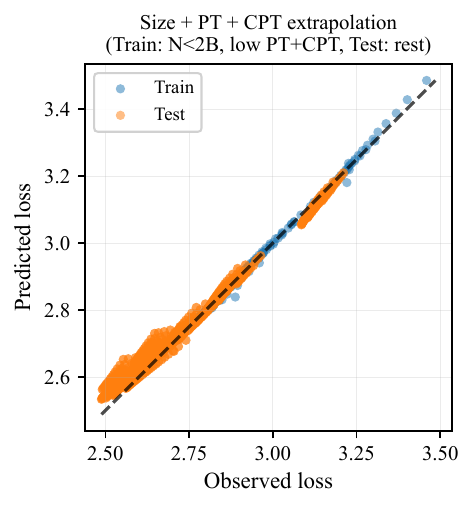}
    \caption{Predicted loss vs. observed loss across multiple scenarios.  The scaling laws are fit on smaller models with varying amount of PT and CPT token budgets.  Loss is estimated over domain 0.}
    \label{fig:scaling_law_loss_observed_pred_0}
\end{figure*}

We train the scaling law in (3) using data from domain $k=0$ in addition to results in Section~\ref{sec:scaling_law_fitting}.  We consider the same three scenarios for testing:
\begin{itemize}
    \item Scenario 1: Train on small model size ($< 2.7$B), and test on larger model size ($2.7$B)
    \item Scenario 2: Train on small model size ($< 2.7$B) and low $D$, test on rest
    \item Scenario 3:  Train on small model size ($< 2.7$B) and low $D$ and low $D_k$, and test on rest.
\end{itemize}

\cref{tab:scaling_law_mre_0} describes our results in terms of the mean absolute error (MAE) and $R^2$ coefficient.  Both MAE is low from $0.6-1.5$, and the $R^2$ is high and close to 1 even for setting where we test on only a small handful of runs - 130 for scenario 3. When testing with similar number of runs but for smaller models we have overall low MAE and high $R^2$. We also plot the observed loss vs. predicted loss in \cref{fig:scaling_law_loss_observed_pred_0}.

\begin{table}[ht]
    \centering
    \begin{tabular}{l|c|c|c|c}
    \hline
         & Train Size & Test Size &  MAE & R2 \\
       \hline
        Scenario 1 & 649 & 142 & 0.009 & 0.934 \\
        Scenario 2 & 344 & 447 & 0.025 & 0.960\\
        Scenario 3 & 156 & 640 & 0.022 & 0.973 \\
        \hline
    \end{tabular}
    \caption{Fitting the scaling laws on domain 0 with multiple held out test conditions.}
    \label{tab:scaling_law_mre_0}
\end{table}

Our findings are similar to results for domain $k=8$.  We see that the scaling law fits the data well with low MAE and MRE, and high $R^2$.

\subsection{Aggregated Results Across All Domains}

We train a scaling law for each individual domain $k$.  We report the average correlation, MAE, and MRE across domains for the 1.3B parameter models.  Data from each of the models starts with PT in $\{10, 20, 40, 80, 120, 160\}$ thousand steps.  Results are summarized in \cref{tab:per_cluster_metrics}.  We see that all clusters have low MAE, MRE, and high correlation indicating a good fit for each cluster.

\begin{table}[t]
\centering
\small
\begin{tabular}{r r r  c c c  c c c}
\toprule
 & & & \multicolumn{3}{c}{Train} & \multicolumn{3}{c}{Test} \\
\cmidrule(lr){4-6} \cmidrule(lr){7-9}
Domain & $n_\text{train}$ & $n_\text{test}$ & MAE & MRE & $r$ & MAE & MRE & $r$ \\
\midrule
    0 & 113 & 60 & 0.0021 & 0.0008 & 0.9993 & 0.0014 & 0.0005 & 0.9959 \\
    1 & 114 & 60 & 0.0023 & 0.0008 & 0.9945 & 0.0024 & 0.0009 & 0.9988 \\
    2 & 113 & 60 & 0.0020 & 0.0008 & 0.9995 & 0.0022 & 0.0009 & 0.9992 \\
    3 & 113 & 60 & 0.0026 & 0.0010 & 0.9989 & 0.0032 & 0.0013 & 0.9965 \\
    4 & 112 & 59 & 0.0020 & 0.0007 & 0.9993 & 0.0014 & 0.0005 & 0.9995 \\
    5 & 114 & 60 & 0.0015 & 0.0005 & 0.9994 & 0.0006 & 0.0002 & 0.9999 \\
    6 & 113 & 60 & 0.0020 & 0.0008 & 0.9995 & 0.0025 & 0.0011 & 0.9996 \\
    7 & 112 & 60 & 0.0016 & 0.0006 & 0.9997 & 0.0013 & 0.0006 & 0.9996 \\
    8 & 113 & 60 & 0.0053 & 0.0019 & 0.9393 & 0.0017 & 0.0008 & 0.9994 \\
    9 & 114 & 59 & 0.0031 & 0.0013 & 0.9954 & 0.0015 & 0.0007 & 0.9993 \\
    10 & 94 & 50 & 0.0020 & 0.0008 & 0.9996 & 0.0017 & 0.0007 & 0.9997 \\
    11 & 112 & 59 & 0.0014 & 0.0005 & 0.9998 & 0.0028 & 0.0010 & 0.9991 \\
    12 & 113 & 51 & 0.0010 & 0.0004 & 0.9999 & 0.0019 & 0.0008 & 0.9988 \\
    13 & 110 & 60 & 0.0016 & 0.0006 & 0.9996 & 0.0027 & 0.0011 & 0.9965 \\
    14 & 111 & 60 & 0.0018 & 0.0006 & 0.9994 & 0.0014 & 0.0005 & 0.9996 \\
    15 & 105 & 49 & 0.0086 & 0.0028 & 0.8697 & 0.0018 & 0.0007 & 0.9985 \\
\midrule
    Avg & & & 0.0026 & 0.0009 & 0.9871 & 0.0019 & 0.0008 & 0.9987 \\
\bottomrule
\end{tabular}
\caption{Per-domain scaling law fit metrics for the 1.3B model. Scaling laws are trained with the CPT steps $< 20$B tokens, and  tested with CPT steps $\geq 20$B tokens.}
\label{tab:per_cluster_metrics}

\end{table}

\section{Zero-shot QA Performance with Large Token Budget}
\label{sec:125M_optimal}
Our results in Section~\ref{sec:exp_lm} use 1.3B parameter models at up to 420B tokens (roughly $14\times$ Chinchilla).  However, we expect the performance to vary with the length of training as a function of both model size and total training budget. In this section, we show performance when scaling the amount of data to large compute budget.  In order to do so, we consider a smaller model size with 125M parameters.  We train for up to 200B tokens, which corresponds to a $100\times$ Chinchilla multiplier.  Although these models are trained on less total data, the relative scale of both model size and data makes them more overtrained, and we expect to see results closer to the overtrained small models in Section~\ref{sec:synthetic}.  Note further that we expect to see similar trends when scaling to larger compute budgets for 1.3B models but is sufficient to test in the smaller setting.

\begin{figure*}[ht]
    \centering
       \begin{subfigure}{0.3\textwidth}
        \centering
        \includegraphics[width=\textwidth]{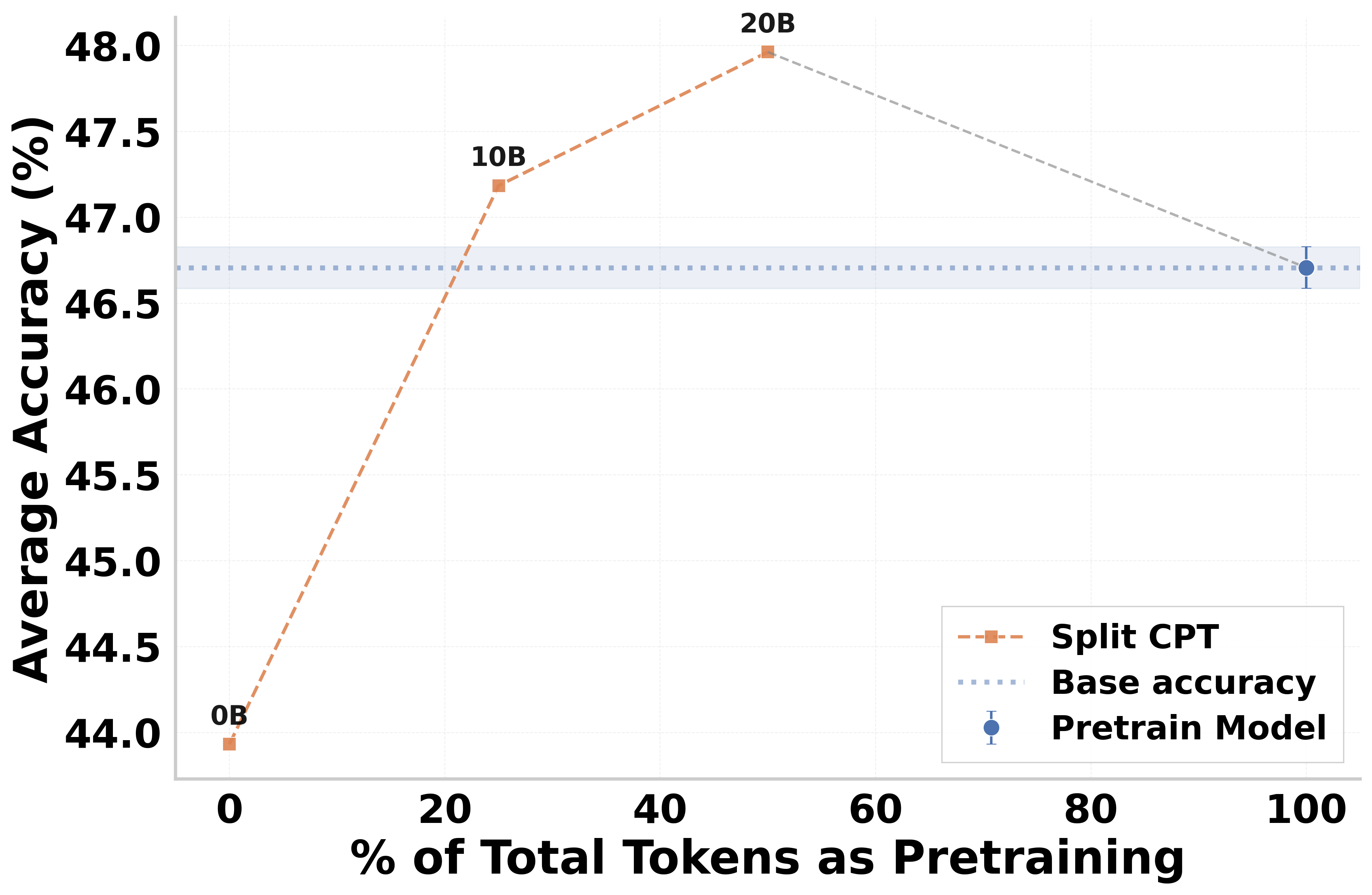}
        \caption{40B}
        \label{fig:125M_40k_pt}
    \end{subfigure}
    \begin{subfigure}{0.3\textwidth}
        \centering
        \includegraphics[width=\textwidth]{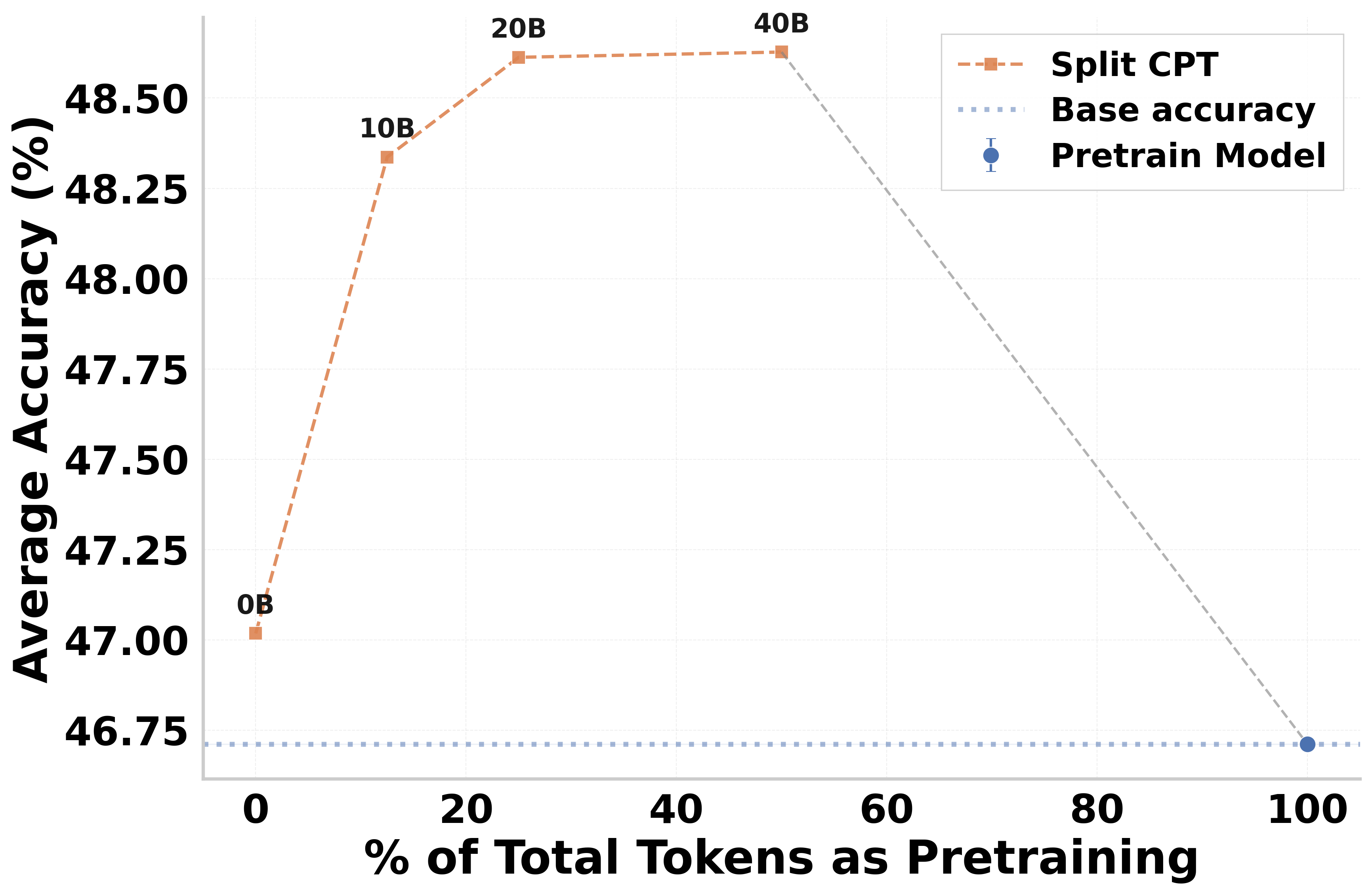}
        \caption{80B}
        \label{fig:125M_80k_pt}
    \end{subfigure}
    \begin{subfigure}{0.3\textwidth}
        \centering
        \includegraphics[width=\textwidth]{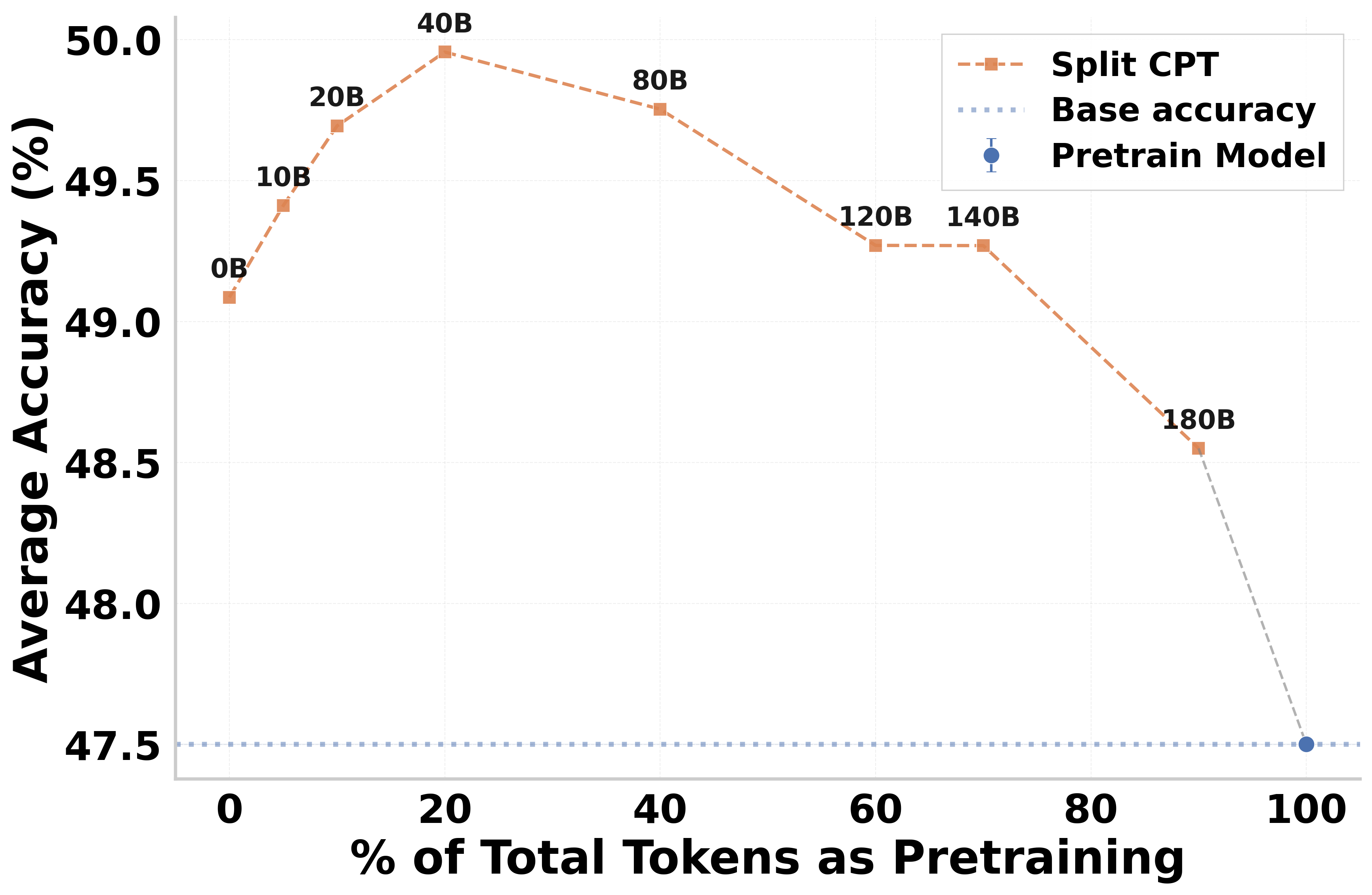}
        \caption{200B}
        \label{fig:125M_200k_pt}
    \end{subfigure}
       
    \caption{Performance across eight zero-shot QA benchmarks for a 125M model at the same training compute with varying amount of pretraining steps. Each subfigure has a different total number of training tokens.}
    \label{fig:zero_shot_qa_125M}
\end{figure*}

Results are presented in \cref{fig:zero_shot_qa_125M}.  We see similar trends as in \cref{fig:phonebook_cpt_short_overlap} where at larger budgets, even the full split training performs better than full pretraining.

\section{1.3B Model Size Zero-shot QA Training Curves}
  Prior experiments focus on performance at the end of training.  However, it is unclear whether the training dynamics of split model training is similar to that of full pretraining.  We compare three sets of models: a pretrained 1.3B model, sixteen 1.3B models with split training on DCLM clusters, and a larger 2.7B model which has twice the number of inference flops.  We choose the three split points: $t_s=40000$, where the model performs similarly to the fully pretrained model,  $t_s=80000$ corresponding to the compute allocation where the model performs similarly to the 2.7B parameter model, and $t_s=160000$ our closest run to the estimated optimal allocation from our scaling law. We see that across all values of $t_s$ the model has similar behavior.  There is a sharp increase for the first few thousand steps of each split model, followed by a similar trend as the fully pretrained model afterwards. 

  \begin{figure*}[ht]
    \centering
       \begin{subfigure}{0.3\textwidth}
        \centering
        \includegraphics[width=\textwidth]{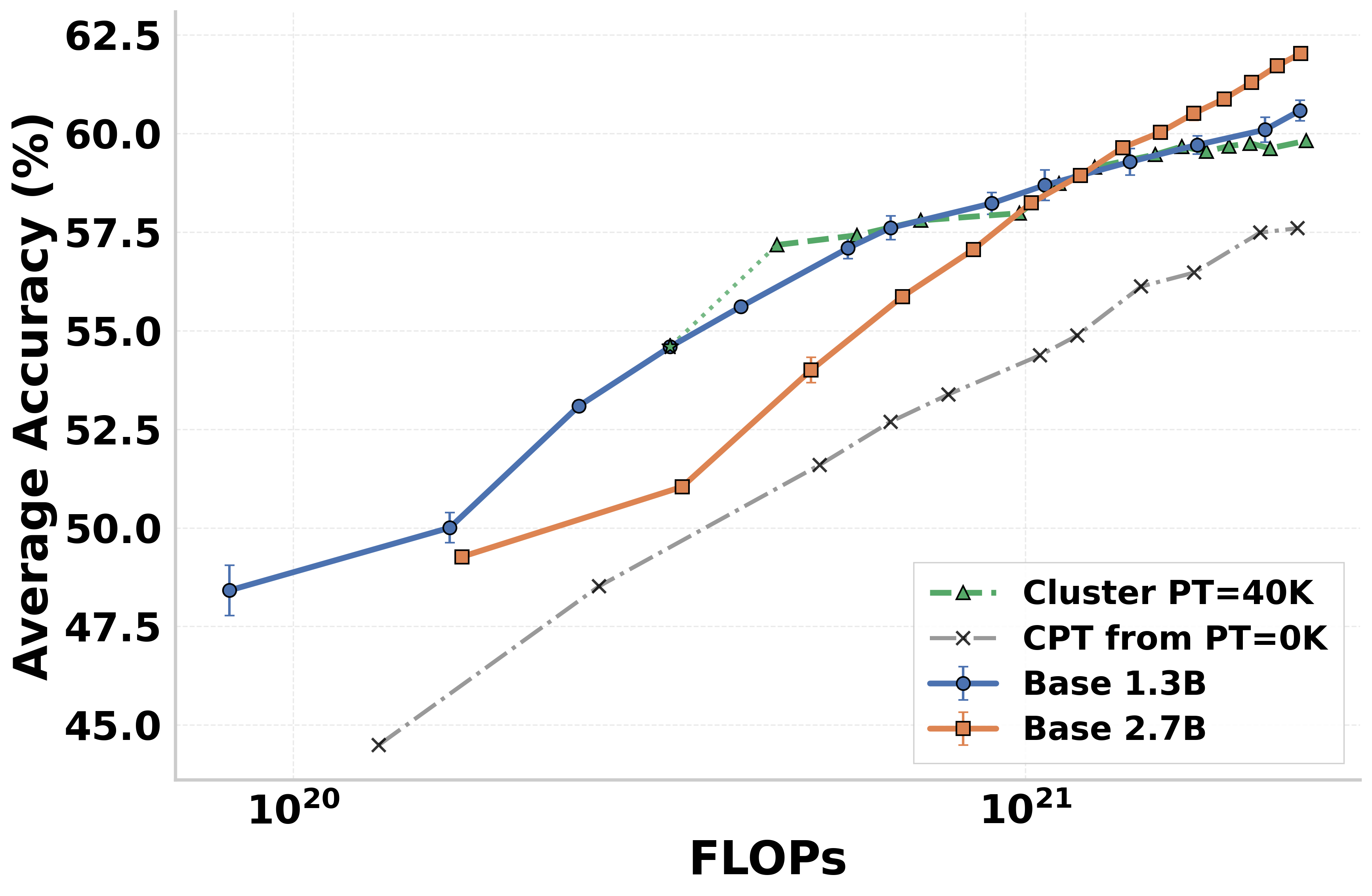}
        \caption{40B PT}
        \label{fig:40k_pt}
    \end{subfigure}
    \begin{subfigure}{0.3\textwidth}
        \centering
        \includegraphics[width=\textwidth]{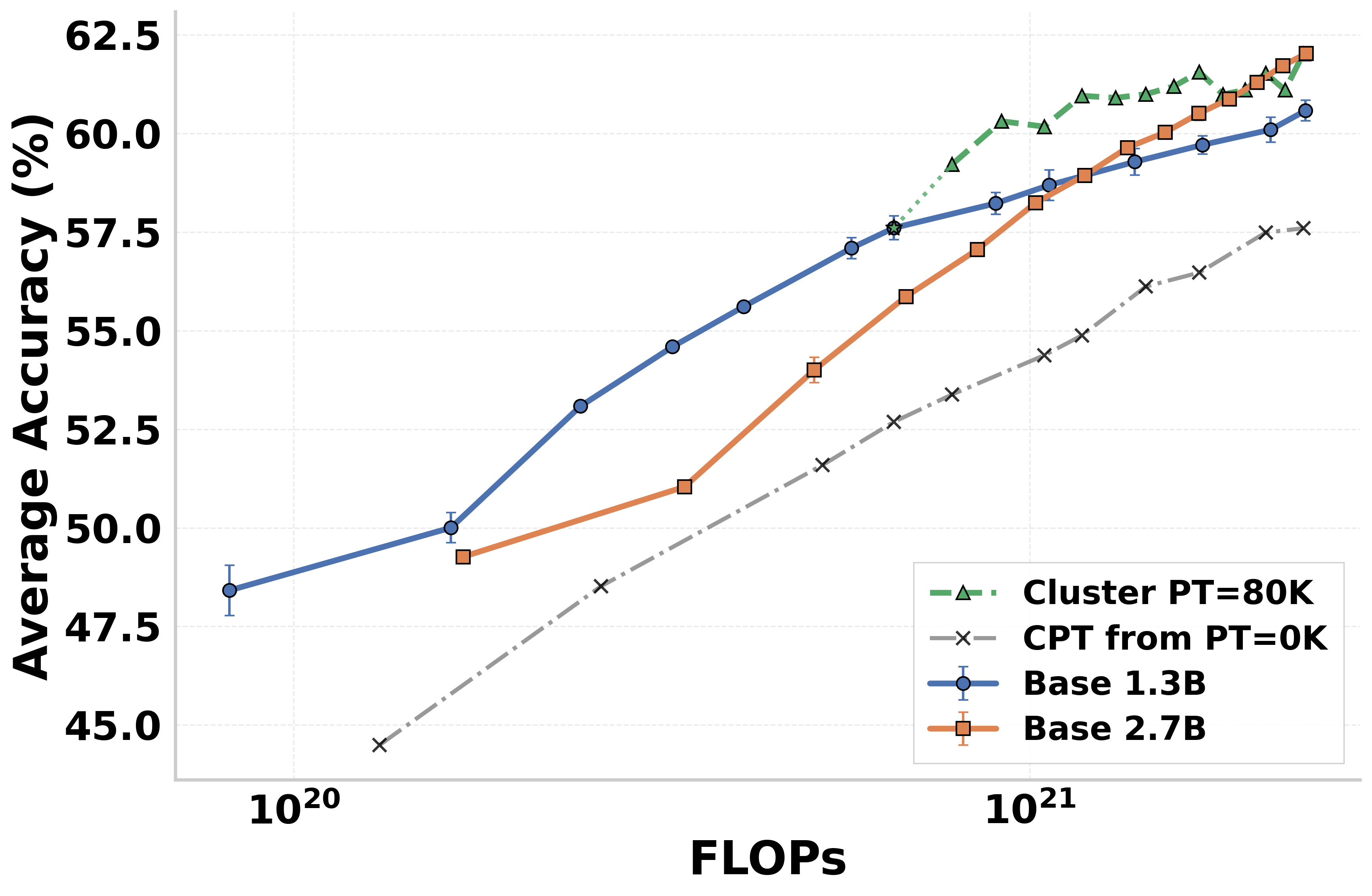}
        \caption{80B PT}
        \label{fig:80k_pt}
    \end{subfigure}
    \begin{subfigure}{0.3\textwidth}
        \centering
        \includegraphics[width=\textwidth]{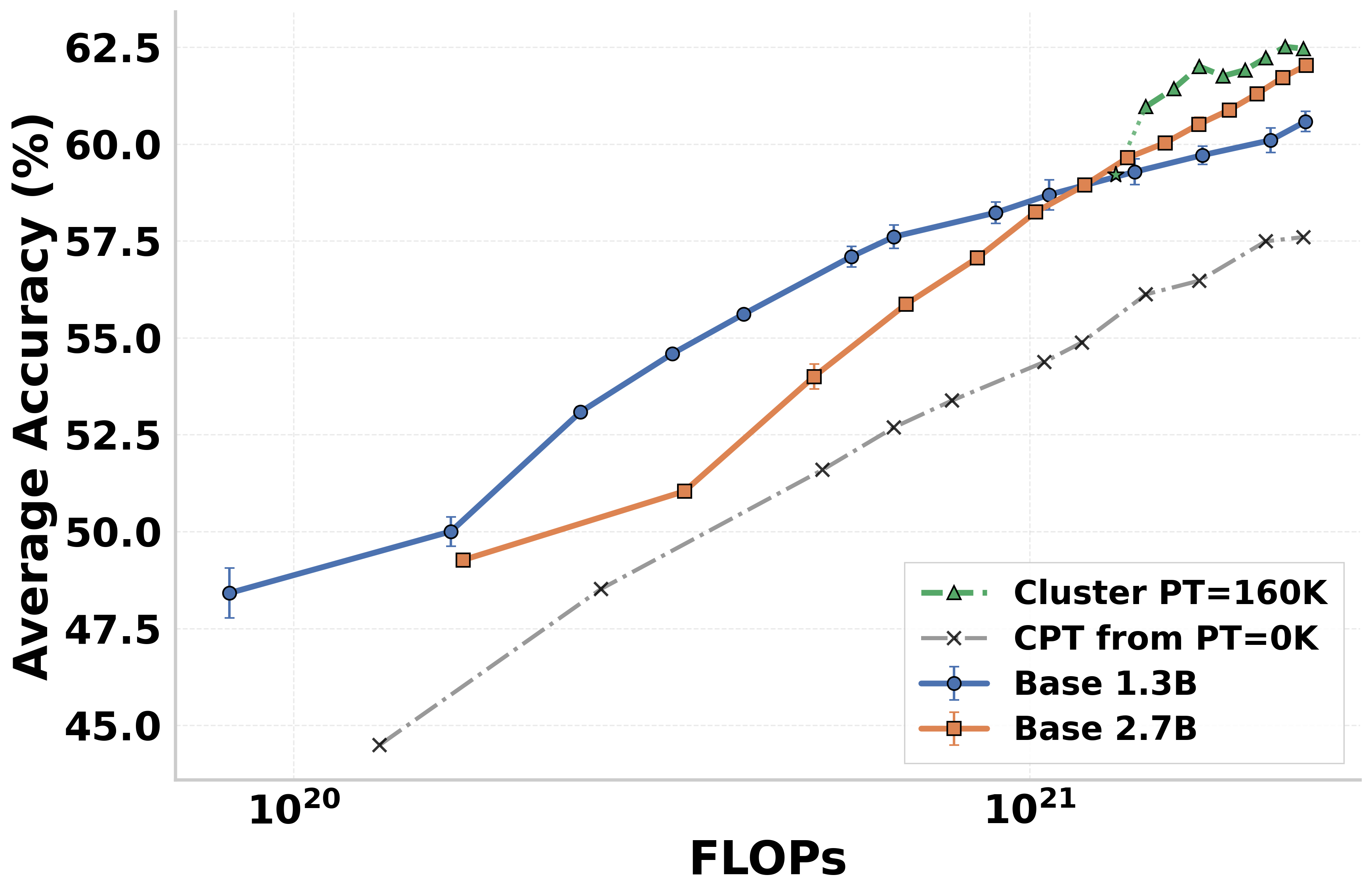}
        \caption{160B PT}
        \label{fig:160k_pt}
    \end{subfigure}       
    \caption{Performance of 1.3B models at the same training compute with different amount of pretraining steps. We compare split training with pretraining from scratch, split training with no pretraining, and finetuning (split training for only 1B tokens).}
    \label{fig:zero_shot_qa}
\end{figure*}

\section{Performance with Varying Number of Domains for 2.7B Models}
\begin{figure}
    \centering
    \includegraphics[width=\linewidth]{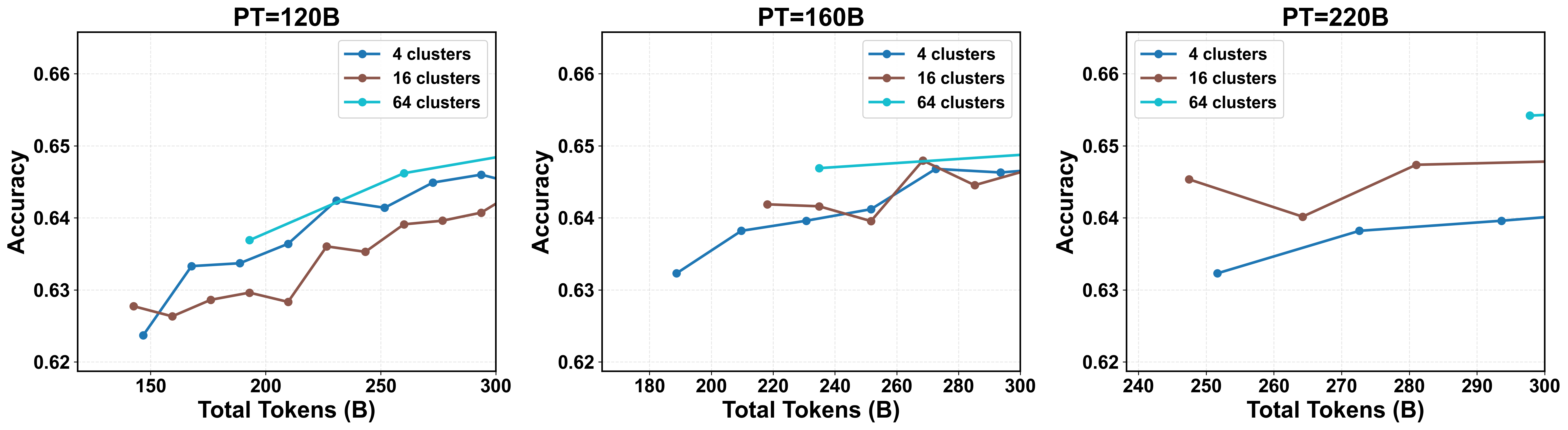}
    \caption{Accuracy vs. number of training tokens for varying number of domains and varying number of PT steps. All domain models are 2.7B parameter models.  The total number of parameters varies with the number of domains, but the number of active parameters remains constant at a single domain model.}
    \label{fig:2_7B_cluster_comparison}
\end{figure}

We extend the experiments for the 1.3B parameter models wiht varying number of domains to the 2.7B setting. Results for the 2.7B parameter model training up to 300B tokens are summarized in \cref{fig:2_7B_cluster_comparison}.  We see that the 64 domain models perform the best at all split training points, although the performance is close to the 4 domain at low $t_s$ and 16 domain at higher $t_s$.  We also see that the performance improvement is relatively flat at later $t_s$ for both the 1.3B and 2.7B indicating that only a small number of steps may be needed for more specialized domains.  In many settings, it can also be expensive to route and maintain 64 domain models.  Thus we find 16 domains to be the best tradeoff of performance at all scales, and number of parameters for practical settings.

\section{Individual Accuracies for Optimal LM Pretraining Allocation}
We report individual task accuracies in \cref{tab:indiv_acc}.

  \begin{table}[htbp]
  \centering
  
  \begin{tabular}{|l|cc|cc|c|}
  \hline
  \textbf{Benchmark} & \textbf{Pretrain 1.3B} & \textbf{Split 1.3B} & \textbf{Pretrain 2.7B} & \textbf{Split 2.7B} & \textbf{Pretrain 7B} \\
  \hline
  ARC-Challenge & 39.25 & \textbf{42.75} & 40.70 & \textbf{45.82} & 42.06 \\
  ARC-Easy      & 68.06 & \textbf{72.31} & 70.62 & \textbf{73.74} & 72.94 \\
  BoolQ         & \textbf{64.01} & 64.19 & 65.78 & \textbf{66.21} & 63.12 \\
  HellaSwag     & 63.91 & \textbf{67.30} & 67.60 & \textbf{69.37} & 69.39 \\
  MMLU          & 35.59 & \textbf{37.04} & 37.42 & \textbf{38.54} & 38.74 \\
  PIQA          & 75.35 & \textbf{78.29} & 78.24 & \textbf{78.29} & 77.75 \\
  SciQ          & 90.00 & \textbf{90.60} & 89.20 & \textbf{89.90} & 91.40 \\
  WinoGrande    & \textbf{61.88} & 62.19 & \textbf{63.85} & 63.22 & 65.19 \\
  \midrule
  \bottomrule
  \end{tabular}
  \caption{QA benchmark accuracy for split model training with optimal token budget allocation.}
  \label{tab:indiv_acc}
  \end{table}

\section{Mixture of Experts Model Perplexity}
\label{sec:moe_ppl}
We test whether our clustering-based split training can be extended with a Mixture-of-Experts (MoE) decoder-only Transformer. We train an MoE model that has 3.8B total parameters and 0.7B active parameters per token, and follows a scaled-down Qwen1.5-style MoE architecture \citep{qwen_moe}.  \cref{tab:moe_cluster_ppl} reports per-cluster perplexities, where the split MoE models are better in every cluster, and have an average relative improvement of 2.2\%.

We report perplexity for each individual cluster for the split Mixture of Experts (MoE) model compared with a single MoE.  We find that improvements from split model training are complementary as training with MoE improves over training a single model.
\begin{table}[t]
\centering
\small
\begin{tabular}{rcc}
\toprule
\textbf{Cluster ID} & \textbf{Base PPL} & \textbf{Split MoE PPL} \\
\midrule
0  & 20.190 & 19.819 \\
1  & 18.075 & 17.731 \\
2  & 18.600 & 18.354 \\
3  & 16.980 & 16.903 \\
4  & 21.036 & 18.414 \\
5  & 19.639 & 19.550 \\
6  & 15.551 & 15.097 \\
7  & 16.774 & 15.972 \\
8  & 15.046 & 14.960 \\
9  & 15.724 & 15.405 \\
10 & 18.052 & 17.783 \\
11 & 22.305 & 22.002 \\
12 & 17.818 & 17.731 \\
13 & 19.544 & 19.461 \\
14 & 20.784 & 20.656 \\
15 & 18.390 & 18.294 \\
\midrule
\textbf{Average} & \textbf{18.407} & \textbf{18.008} \\
\bottomrule
\end{tabular}
\caption{Per-cluster perplexity (PPL) comparing split training of Mixture of Expert (MoE) models with a single model pretrained on all clusters.}
\label{tab:moe_cluster_ppl}
\end{table}

\end{document}